\documentclass{article}

\PassOptionsToPackage{numbers}{natbib}
\usepackage[final]{neurips_2022}

\usepackage{graphicx}
\usepackage{wrapfig}
\usepackage[utf8]{inputenc} 
\usepackage[T1]{fontenc}    
\usepackage[pagebackref,breaklinks,colorlinks]{hyperref}       
\usepackage{url}            
\usepackage{booktabs}       
\usepackage{multirow}
\usepackage{amsfonts}       
\usepackage{nicefrac}       
\usepackage{microtype}      
\usepackage{xcolor}         
\usepackage{amsmath}        
\usepackage{textcomp, gensymb}
\usepackage[ruled,vlined]{algorithm2e}
\usepackage{caption}
\usepackage{enumitem, multirow, xspace}

\usepackage{pifont}
\newcommand{\cmark}{\ding{51}}%
\newcommand{\xmark}{\ding{55}}%
\newcommand{\algname}{TCP\xspace}

\newcommand\blankfootnote[1]{%
  \let\thefootnote\relax\footnotetext{#1}%
  \let\thefootnote\svthefootnote%
}

\title{
\algname: A New Autonomous Driving Paradigm of
Learning Control from Trajectory Planning
}

\title{
On Synergizing Trajectory and Control:\\ a Heterogeneous Multi-task Learning approach for End-to-end Autonomous Driving Planning
}
\title{
When Trajectory Planning Meets Direct Control: a Heterogeneous Mutual Learning Approach for End-to-end Autonomous Driving
}
\title{
Trajectory-guided Control Prediction for End-to-end Autonomous Driving: A Simple yet Strong Baseline
}
\author{%
  Penghao~Wu{$^{\ast\dagger}$} \\
  Shanghai AI Laboratory\\
  Shanghai Jiao Tong University \\
  \texttt{wupenghaocraig@sjtu.edu.cn} \\
  \And
  Xiaosong~Jia{$^\ast$} \\
  Shanghai Jiao Tong University \\
   Shanghai AI Laboratory\\
  \texttt{jiaxiaosong@sjtu.edu.cn}
  \And 
  Li~Chen{$^\ast$} \\
  Shanghai AI Laboratory \\
  \texttt{lichen@pjlab.org.cn} \\
  \And 
  Junchi~Yan{$^\dagger$} \\
  Shanghai Jiao Tong University \\
  Shanghai AI Laboratory \\
  \texttt{yanjunchi@sjtu.edu.cn} \\
  \And 
  Hongyang~Li \\
  Shanghai AI Laboratory \\
  Shanghai Jiao Tong University \\
  \texttt{lihongyang@pjlab.org.cn}
  \And 
  Yu~Qiao \\
  Shanghai AI Laboratory \\
  \texttt{qiaoyu@pjlab.org.cn}

}

\begin{document}

\maketitle

\blankfootnote{$^\ast$ Equal Contribution. Work done when PW and XJ were interns at Shanghai AI Laboratory.} 
\blankfootnote{$^\dagger$ Correspondence author.}

\begin{abstract}
  Current end-to-end autonomous driving methods either run a controller based on a planned trajectory or perform control prediction directly, which have spanned two separately studied lines of research. Seeing their potential mutual benefits to each other, this paper takes the initiative to explore the combination of these two well-developed worlds. Specifically, our integrated approach has two branches for trajectory planning and direct control, respectively. The trajectory branch predicts the future trajectory, while the control branch involves a novel multi-step prediction scheme such that the relationship between current actions and future states can be reasoned. 
  The two branches are connected so that the control branch receives corresponding guidance from the trajectory branch at each time step. The outputs from two branches are then fused to achieve complementary advantages. Our results are evaluated in the closed-loop urban driving setting with challenging scenarios using the CARLA simulator. Even with a monocular camera input, the proposed approach ranks \textit{first} 
  on the official CARLA Leaderboard, outperforming other complex candidates with multiple sensors or fusion mechanisms by a large margin. 
  The source code is publicly available at \url{https://github.com/OpenPerceptionX/TCP}.
\end{abstract}
\section{Introduction} \label{sec:intro}
End-to-end autonomous driving methods, which directly map raw sensor data to a planned trajectory or low-level control actions, show the virtue of simplicity, conceptually avoiding the cascading error of complex modular design and heavy hand-crafted rules.
The output prediction of the model for end-to-end autonomous driving generally falls into two forms: trajectory/waypoints \citep{rhinehart2019deep, pilotnet2020nvidia, chen2020lbc, prakash2021transfuser, chitta2021neat, jaeger2021transfuser+, chen2022lav} and direct control actions \citep{codevilla2018cil, liang2018cirl, codevilla2019cilrs, ohn2020lsd, chen2021wor, zhang2021roach, chekroun2021gri}. However, there is still no clear conclusion as to which of these two forms is better for all circumstances or certain scenarios.

\begin{figure}[t!]
      \centering
      \includegraphics[width=\textwidth]{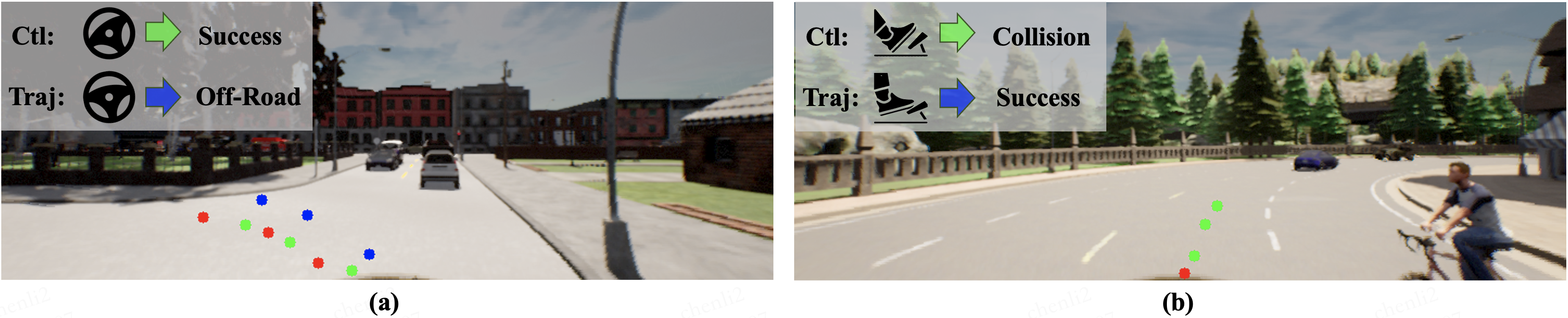}
      \caption{Typical failure cases of two prediction paradigms. \textcolor{red}{Red} dots indicate the trajectory prediction, \textcolor{blue}{blue} dots are the actual path following the \textcolor{red}{red} trajectory with PID controllers, and \textcolor{green}{green} dots denote the actual path from control-based method. (a)  trajectory-based methods may struggle for big turns. (b) control-based methods may have a reaction latency and suffer from abrupt obstacles due to focusing on current time step only. 
      These observations motivate us to propose a unified framework to combine these two worlds for mutual benefits.
      }
    \label{fig:motivation}
  \end{figure}

Different from control predictions that could be directly applied to the vehicle, for methods that \textbf{plan trajectory}, additional controllers such as PID controllers are usually needed as a subsequent step to convert the planned trajectory into control signals.
One attractive and potential supremacy of trajectory-based prediction is that it actually considers a relatively longer time horizon into the future and could be further combined with other modules (\textit{e.g.}, multi-agent trajectory prediction \citep{zhang2021lbw, chen2022lav}, semantic or occupancy prediction modules \citep{chitta2021neat, casas2021mp3, hu2022stp3}) to reduce possible collisions.
However, turning the trajectory into control actions so that the vehicle could follow the planned trajectory is not trivial \citep{zablocki2021xai}. The industry usually adopts sophisticated control algorithms such as model predictive control to achieve reliable trajectory-following performance \citep{camacho2013mpc, guo2019mpcfollowing}. Simple PID controllers may perform worse in situations such as taking a big turn or starting at the red light due to the inertial problem of end-to-end models \citep{jaeger2021transfuser+}.
For \textbf{control-based} methods, the control signals are directly optimized. Nevertheless, their focus on the current step may cause deferred reactions to avoid potential collisions with other moving agents. The independence between the control predictions of different steps also makes the actions of the vehicle more unstable or discontinuous.
Fig.~\ref{fig:motivation} shows two typical cases where two paradigms fail respectively.
How to combine these two forms of prediction model as well as their outputs is an interesting yet relatively rarely studied area, which motivates this work. 

One straightforward (but in fact rarely studied in literature) idea is to train a control prediction model and a trajectory planning model separately, and combine their ultimate outputs directly. It can be viewed as an ensemble of two different models.
However, such a naive approach not only doubles the size of the model, but also ignores possible useful correlations between these two forms. To this end, we introduce the \textbf{\algname} (Trajectory-guided Control Prediction) framework, packing these two branches into a unified framework.
It can be viewed as a multi-task learning (MTL)~\cite{caruana1997mtl, argyriou2006mtfl} framework where a shared backbone extracts common features with decreased computational complexity 
as well as the increased ability of generalization due to the close relationship between the two tasks \citep{liang2019multi3d, kumar2021omnidet, chen2022persformer}.
Furthermore, to address the drawbacks of current control prediction methods, we delicately devise a novel multi-step control branch and a trajectory-guided control prediction scheme.

%
While trajectory planning considers several steps into the future, directly learning the control in a behavior cloning fashion \citep{pomerleau1988alvinn, muller2005e2eavoidance, codevilla2018cil, codevilla2019cilrs, chen2020lbc} often focuses on the current time step only, given prior on each state-action pair as independent and identical distributed (IID).
This assumption is not accurate and may hamper the long-term performance since the driving task is a sequential decision-making problem. To alleviate the problem, we propose  to predict multi-step control actions into the future.
However, the multi-step control process needs interactions with the environment. Thus we formulate a temporal module to learn the forward process and interactions between the ego agent and the environment.
A temporal module implemented with GRU \citep{GRU} progressively deals with the feature representation for each time step,  implicitly taking into account the dynamic motion of agents, interaction among them and dynamic environment information such as the changing of traffic lights.
%
%
%
%

Additionally, to generate accurate control signals in the multi-step prediction scheme, the model should retrieve proper location information from current sensor input for different future time steps. For example, an agent may pay more attention to nearby regions for a few early future time steps and far away regions for the remote ones. Considering that the knowledge has already been partly encoded in the trajectory branch, we adopt the attention mechanism to locate those critical and helpful areas in the long-term trajectory prediction branch, and guide the control prediction branch to pay attention to them at each future step in a corresponding way.
As a result, our model is capable of reasoning about how to optimize current control prediction so that the future states are similar to those from the expert when the predicted control actions are applied.

With the predicted trajectory and control signals from two branches, we propose a situation based fusion scheme to adaptively combine these two forms in a self-ensemble way to form the ultimate output according to the experiments results and prior knowledge.
%
%
It combines the best of these two forms, which further boosts the performance under different scenarios.

\algname has shown superior performance when being validated in the CARLA driving simulator \citep{Dosovitskiy17carla}. Our method, which only uses a \textbf{monocular camera}, achieves a \textbf{75.137} driving score and ranks \textbf{1st} on the public CARLA Leaderboard \citep{carlaleaderboard}, even surpassing prior state-of-the-art methods using multiple cameras and a LiDAR by 13.291 points. The main \textbf{contributions} of this paper include:
\begin{itemize}
\item We examine two dominant paradigms for end-to-end autonomous driving: trajectory planning and direct control, and propose to combine them in an integrated learning pipeline. To our knowledge, this is the first time that such two branches are jointly learned and fused for prediction.


\item 
A multi-step control prediction branch with a temporal module and trajectory-guided attention is devised to enable temporal reasoning. To combine the best of two branches, we design a situation based scheme to fuse the two outputs. 


\item As a simple yet strong baseline, our method with only a monocular camera as input achieves new state-of-the-art on the CARLA Leaderboard with many competitors using multiple sensors. We conduct thorough ablation studies to verify the effectiveness of our approach.
\end{itemize}

\section{Related Work} \label{sec:related_work}
\subsection{End-to-end Autonomous Driving}


Learning-based end-to-end autonomous driving has emerged as an active research topic in recent years. Studies usually fall into two categories: reinforcement learning (RL) and imitation learning.
RL is a promising way to address the problem of being more robust to the distribution shifts of datasets. Liang \textit{et al.} \citep{liang2018cirl} use DDPG to train a policy which is pre-trained in a supervised way. Kendall \textit{et al.} \citep{kendall2019driveaday} train their deep RL algorithm onboard to efficiently learn to drive a real-world vehicle.
The perception task is decoupled out of the online RL process in \citep{toromanoff2020modelfreerl, chekroun2021gri, zhao2022cadre}. The model-based method WoR \citep{chen2021wor} assumes world on rails and uses policy distillation to realize powerful performance.

Imitation learning (IL), especially behavior cloning, collects recorded data for models to mimic with high data efficiency. The expert data typically has two forms, trajectories and control actions.
Zeng \textit{et al.} \citep{zeng2019nmp} train a cost volume to generate the planning route, while \citep{sadat2020p3, casas2021mp3, hu2022stp3} explicitly design safety and comfort costs based on semantic occupancy maps to select the best one in the expert trajectory sets.
Zhang \textit{et al.} \citep{zhang2021lbw} predict trajectories of surrounding vehicles with labeled BEV map. LBC \citep{chen2020lbc} and NEAT \citep{chitta2021neat} decode waypoints from a dense heatmap or offset map. 
These approaches aforementioned all utilize a relatively dense representation to obtain results which increases model complexity. Transfuser and its variants \citep{prakash2021transfuser, jaeger2021transfuser+} adopt a simple GRU to auto-regress waypoints. 
LAV \citep{chen2022lav} adopts a temporal GRU module to further refine the trajectory. They unanimously achieve impressive performance on the CARLA leaderboard, motivating us to adapt the auto-regression scheme as well in our design. 
On the other hand, 
all trajectory-based methods use PID controllers to get the ultimate actions, which may cause inferior effects in complicated scenarios. 

Another genre to  predict control actions directly is 
proposed in \citep{pomerleau1988alvinn, muller2005e2eavoid, bojarski2016nvend, xu2017lstme2e}. 
%
CIL \citep{codevilla2018cil} adds a measurement encoder and multiple branches for different high level commands with the image encoder.
CILRS \citep{codevilla2019cilrs} is proposed afterwards and further introduces a speed prediction head. They stand as classic baselines for IL in the CARLA driving simulator. Diverse optimized approaches are presented based on them, such as multi-modal inputs \citep{hawke2020urbancil, xiao2020multimodal}, multi-task learning \citep{yang2018e2emmmt, li2018rethinking, hou2019fmnet, ishihara2021multie2eattention, kim2020fasnet, huch2021e2ecav, zhu2022mtcil}, dataset aggregation \citep{prakash2020darb} and knowledge distillation \citep{zhao2021sam, zhang2021roach}.
However, the compact control-based methods often have higher vehicles collision rates, remaining an interesting domain to explore.
%
%
Similar work exists in other related domains such as robotic navigation as well. \citep{pokle2019robot} learns a controller after a local trajectory planner to improve the overall navigation behavior.

\subsection{Multi-task and Ensemble Learning for Autonomous Driving}



Multi-task learning is a popular approach to train several related tasks simultaneously to help each other and improve generalization \citep{caruana1997mtl, argyriou2006mtfl}. Combinations of various autonomous driving tasks such as object detection, lane detection, semantic segmentation, depth estimation, \textit{etc}. have been proved to be capable of achieving incredible performance \citep{liang2019multi3d, chennupati2019multinet++, rajamanoharan2019mtmlreid, kumar2021omnidet, wu2021yolop, chen2022persformer}.
MTL is also suitable in the end-to-end problem since it is observed the performance of a direct mapping from an image to control signals is limited.
\citep{yang2018e2emmmt} adds a speed prediction task similar to CIL \citep{codevilla2018cil} and \citep{zhu2022mtcil} separates the lateral and longitudinal controls as two tasks. 
LAV \citep{chen2022lav} trains an extra scene mapping network, and \citep{hou2019fmnet, ishihara2021multie2eattention, jaeger2021transfuser+} additionally predict optical flow or dense depth.
Our idea of training trajectory and control simultaneously is closely related to FASNet \citep{kim2020fasnet}. FASNet predicts future positions of the ego agent as an auxiliary task and adds a kinematic loss considering the relation between control and locations. However, the constrain is based on a constant velocity model which neglects the important throttle and brake, and it does not work at the inference time. On the other hand, our \algname framework has feature interactions at an earlier stage to fully explore their potential mutual benefits.
%
%
%

Ensembles of models have long been utilized to improve the performance in computer vision \citep{dietterich2000ensemble, krahenbuhl2015lpo, lakshminarayanan2017deepensemble, vyas2018oodensemble, xu2020mmensemble, prakash2020darb}.
Besides the normal combination of models, two classic ensemble learning methods are particularly preferred in the autonomous driving regime. One is the Test-Time Augmentation (TTA), which is of great help to the 3D object detection task with LiDAR \citep{carranza2021odadensemble, li2022deepfusion}.
%
%
Another one is the fusion of experts \citep{jacobs1991expertmixture} where experts are trained on a subset of the input space and a gating network is trained to provide the fusion weights. LSD \citep{ohn2020lsd} and MoDE \citep{kim2022mode} divide a dataset into sub-scenarios to get different sub-policies for end-to-end autonomous driving.
%
%
These traditional ensemble approaches combine models of the same structure while our approach tries to combine two different representations.
Also, the multiple experts design increases the complexity of the training strategy and we seek to have a simpler situation based fusion scheme to boost the performance.

\section{Trajectory-guided Control Prediction} \label{sec:method}
\subsection{Problem Setting} \label{sec:prob_set}

\textbf{Problem formulation.} Given the state $\rm{\bf {x}}$ comprised of the sensor signal $\rm{\bf {i}}$, the speed of the vehicle $v$, and the high level navigation information $\rm{\bf g}$ including a discrete navigation command and the coordinates of navigation target provided by the global planner, the end-to-end model needs to output control signals $\rm{\bf a}$ comprised of longitudinal control signals \textit{throttle} $\in [0, 1]$ and \textit{brake} $\in [0, 1]$, and the lateral control signal \textit{steer} $\in [-1, 1]$.

Conventional methods tackle this problem with either a trajectory-output or a control-output only model. However, \algname combines both of them as two branches: a trajectory branch which predicts the planned trajectory and a control branch which is guided by the trajectory one and outputs both current and multi-step control signals into the future. Both branches are trained in a supervised manner.
Consider an expert which directly outputs the control signals at each step, supervising the predicted trajectory with the ground truth trajectory makes it not strictly satisfy the setting of behavior cloning in imitation learning. The ground truth trajectory indeed involves future expert actions and future states about the environment, so we formulate it as a trajectory planning task with ground truth trajectory as supervision for our trajectory branch. As for the control branch, training a control model which 
makes current control prediction supervised by the expert control is just behavior cloning in imitation learning, and it can be formulated as:
\begin{equation}
    \label{eqn:singlestepobjective}
   \arg \min_{\theta} \mathbb{E}_{(\rm{\bf {x}}, \rm{\bf {a}^*})\sim D} [\mathcal{L}( \rm{\bf {a}^*}, \pi_\theta(\rm{\bf {x}}))],
\end{equation}
where $D = \{(\rm{\bf {x}}, \rm{\bf {a}^*})\}$ is a dataset comprised of state-action pairs collected from the expert. $\pi_\theta$ denotes the policy of the control branch, and $\mathcal{L}$ is the loss  measuring how close the action from the expert and the action from our model is. The expert collects the dataset by controlling the vehicle and interacting with the world. Each collected route is a trajectory $\xi = (\rm{\bf {x}}_0, \rm{\bf {a}}^*_0, \rm{\bf {x}}_1, \rm{\bf {a}}^*_1, \cdots , \rm{\bf {x}}_T)$ as a sequence of state action pairs $\{(\rm{\bf {x}}_i, \rm{\bf {a}}^*_i)\}_{i=0}^T$, which is then added into the whole dataset $D$. 

\textbf{Expert demonstration.}
 Here we choose Roach \citep{zhang2021roach} as the expert. Roach is a simple model trained by RL with privileged information, including roads, lanes, routes, vehicles, pedestrians, traffic lights, and stops, all being rendered into a 2D BEV image. Such a learning-based expert can transfer more information besides the direct supervision signals compared with an expert made by hand-crafted rules.
Specifically, we have a feature loss which forces the latent features before the final output head from the student model to be similar to that of the expert. A value loss is also added as an auxiliary task for the student model to predict an expected return.

\begin{figure}[t!]
    \centering
    \includegraphics[width=0.85\textwidth]{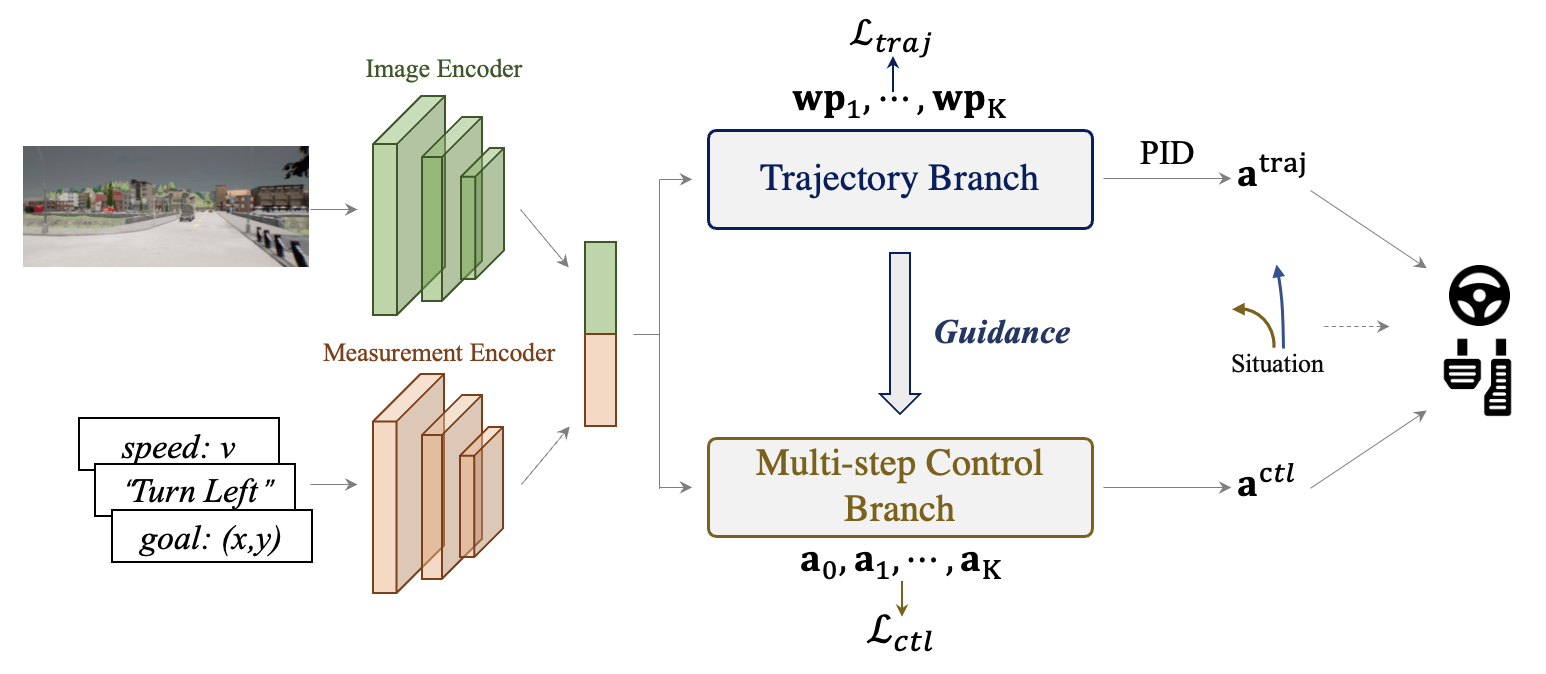}
    \caption{Overview of Trajectory-guided Control Prediction (\algname).
    The encoded features are shared by the trajectory and multi-step control branch. 
    %
    The trajectory branch provides per-step guidance for multi-step control prediction. Outputs from two branches are combined 
    according to our situation based fusion scheme 
    to generate the ultimate control actions. 
    }
    \label{fig:pipeline}
\end{figure}

\subsection{Architecture Design} \label{sec:arch}
\textbf{Overview.} As illustrated in Fig.~\ref{fig:pipeline}, the whole architecture is comprised of an input encoding stage and two subsequent branches.
The input image $\rm{\bf {i}}$ goes through a CNN based image encoder, such as ResNet \citep{he2016resnet}, to generate a feature map $\rm{\bf F}$.
In the meantime, the navigation information $\rm{\bf g}$ is concatenated with the current speed $v$ to form the measurement input $\rm{\bf m}$, then an MLP based measurement encoder takes $\rm{\bf m}$ as its input and outputs the measurement feature $\rm{\bf j}_m$.
The encoded features are then shared by two branches for subsequent trajectory and control predictions. Specifically, the control branch is a novel multi-step prediction design with guidance from the trajectory one, which will be illustrated in detail in the following sections.
Finally, a situation based fusion scheme is adopted to combine the best of the two output paradigms.
We will go over each part in detail below.

\subsubsection{Trajectory planning branch} \label{sec:traj_branch}
Different from control prediction which directly predicts control actions, the trajectory planning branch first generates a planned trajectory comprised of waypoints at $K$ steps for the agent to follow, and then the trajectory is processed by low-level controllers to get the final control actions.
%
%
With the shared feature from the input encoder, the image feature map $\rm{\bf F}$ is average pooled and concatenated with the measurement feature $\rm{\bf j}_m$ to form $\rm{\bf j}^{traj}$. Inspired by \citep{prakash2021transfuser}, we feed $\rm{\bf j}^{traj}$ into a GRU \citep{GRU} to auto-regressively obtain future waypoints one by one to form the planned trajectory altogether. 

We have two PID controllers for longitudinal and lateral control respectively. With the planned trajectory, we first calculate the vectors between consecutive waypoints. The magnitudes of these vectors represent the desired speed and are sent to the longitudinal controller to generate $throttle$ and $brake$ control actions, and the orientations are sent to the lateral controller to get the $steer$ action.

\subsubsection{Multi-step control prediction branch} \label{sec:ctl_branch}
As discussed in Sec.~\ref{sec:prob_set}, for a control model predicting current control actions based on current input only, the supervised training is just behavior cloning, which relies on the independent and identically distributed (IID) assumption.
This assumption apparently does not hold because of the distribution shifts in test cases, since the closed-loop tests require sequential decision making where the historical actions will affect the future states and actions.
Instead of modeling it as a Markov Decision Process (MDP) and resorting to reinforcement learning, here we devise a simple way to mitigate the problem by predicting multi-step control into the future. 

Given the current state $\rm{\bf {x}}_t$, now our multi-step control prediction branch outputs multiple actions: $\pi_{\theta_{multi}} = (\rm{\bf {a}}_t, \rm{\bf {a}}_{t+1}, \cdots, \rm{\bf {a}}_{t+K})$.
However, it is difficult to predict future control actions since we only have sensor inputs at the current time step. Towards this problem, we devise a temporal module to implicitly carry out the changing and interaction process of the environment and our agent. It is supposed to provide mainly dynamic information about the environment and the status of the agent itself, such as the motion of other objects, the changing of traffic lights, and the status of the ego agent.
Meanwhile, to improve the ability of incorporating critical static information (\textit{e.g.}, curbs and lanes) and boost the spatial consistency of two branches,
%
%
we propose to use the trajectory branch to guide the control counterpart to attend to proper regions of the input image at each future time step.

\textbf{Temporal module.} 
Our temporal module is implemented with a GRU for better consistency with the trajectory branch.
At step $t\,(0 \leq t \leq K-1)$, the input  for the temporal module is the concatenation of the current feature $\rm{\bf j}^{ctl}_{t}$ (more construction details in the next section) and current predicted action $\rm{\bf a}_t$, which is a compact representation about the current states of the environment and the agent itself.
The temporal module is supposed to reason about the dynamic changing process based on current feature vector and the predicted action.
Then the updated hidden state $\rm{\bf h}^{ctl}_{t+1}$ will contain dynamic information about the environment and the updated status of the agent at time step $t+1$.
To some extent, the temporal module acts as a coarse simulator with the whole environment and the agent being abstracted as a feature vector. It then simulates the interaction between the environment and the agent based on current prediction of actions.


\textbf{Trajectory-guided attention.} 
With the sensor input at current step only, it is hard to pick out desirable regions where the model should focus on at future steps. However, the location of the ego agent contains important cues about how to find those regions containing critical static information for control prediction at each step.

\setlength{\intextsep}{3pt}
\captionsetup[wrapfigure]{font=footnotesize}
\begin{wrapfigure}{r}{0pt}
\centering
        \includegraphics[width=0.54\textwidth]{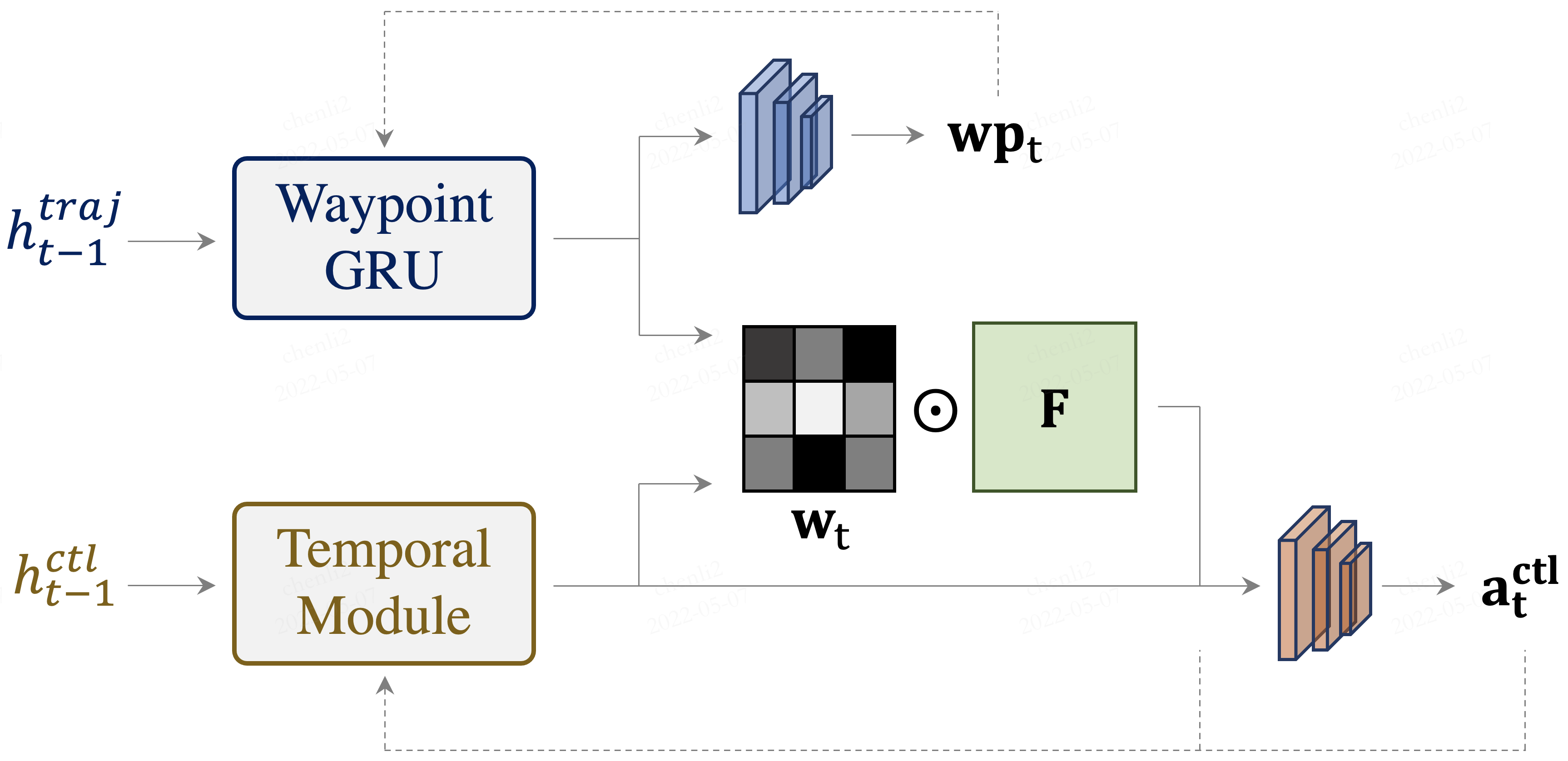}
    \caption{Detailed trajectory guiding process. For predictions at time step $t$, the hidden states from the waypoint GRU and the temporal module are combined to learn an attention weight map to re-aggregate the 2D image feature map for  control prediction.}
    \label{fig:traj_att}
\end{wrapfigure}

Therefore, we seek help from the trajectory planning branch to get information about the possible location of our agent at that corresponding step. As shown in Fig.~\ref{fig:traj_att}, \algname implements this by learning an attention map to extract important information from the encoded feature map. 
The interaction between two branches enhances the consistency of these two strongly related output paradigms and further elaborates the multi-task spirit.
Specifically, with the 2D feature map extracted by the image encoder $\rm{\bf F}$ at time step $t\,(1 \leq t \leq K)$, we calculate an attention map $\rm{\bf w}_t \in \mathbb{R}^{1\times H\times W}$ using the corresponding hidden states from the control branch and the trajectory branch:
\begin{equation}
    \rm{\bf w}_t = \rm{MLP}(\rm{Concat[{\bf h}^{traj}_t, {\bf h}^{ctl}_t]}).
\end{equation}
The attention map $\rm{\bf w}_t \in \mathbb{R}^{1\times H\times W}$ is adopted to aggregate the feature map $\rm{\bf F}$ for this step.
We then combine the attended feature map with $\rm{\bf h}^{ctl}_{t}$ to form the informative representation feature $\rm{\bf j}^{ctl}_{t}$ containing both static and dynamic information about the environment and the ego agent. The process can be described as follows:
\begin{equation}
    \rm{\bf j}^{ctl}_{t} = \rm{MLP}(\rm{Concat[Sum(Softmax(\rm{\bf w}_t) \odot \rm{\bf F}), {\bf h}^{ctl}_t]}).
\end{equation}
The informative representation feature $\rm{\bf j}^{ctl}_{t}$ is fed into a policy head which is shared among all time steps to predict the corresponding control action $\rm{\bf a}_t$. 
Note that for the initial step, we only use the measurement feature to calculate the initial attention map and combine the attended image feature with the measurement feature to form the initial feature vector $\rm{\bf j}^{ctl}_{0}$. To guarantee the feature $\rm{\bf j}^{ctl}_{t}$ does describe the state at that step and contain the important information for control prediction, we add a feature loss at each step to make $\rm{\bf j}^{ctl}_{t}$ close to the feature of the expert as well. 

To this end, our \algname framework endows the model with the reasoning ability along a short time horizon. It emphasizes how to make current control prediction close to the one from the expert. Furthermore, it takes into account what current control prediction can make the environment states and status of ego agent in future time steps similar to the ones from the expert.

\subsection{Loss Design} \label{sec:loss}
Our loss contains trajectory planning loss $\mathcal{L}_{traj}$, control prediction $\mathcal{L}_{ctl}$, and auxiliary loss $\mathcal{L}_{aux}$.

For the trajectory planning branch, the loss $\mathcal{L}_{traj}$ can be expressed as:
\begin{equation}
    \mathcal{L}_{traj} = \sum_{t=1}^K\|\rm{\bf wp}_t - \rm\hat{{\bf wp}_t}\|_1 + \lambda_F \cdot \mathcal{L}_F\left(\rm{\bf j}^{traj}_0, \rm{\bf j}^{Expert}_0\right),
\end{equation}
where $\rm{\bf wp}_t, \rm\hat{{\bf wp}_t}$ are the predicted and ground truth waypoint at the $t^{th}$ step respectively.
 $\mathcal{L}_F$ indicates the feature loss measuring the $L_2$ distance between $\rm{\bf j}^{traj}_0$ and the feature $\rm{\bf j}^{Expert}_0$ from the expert at the current step as an additional supervision signal~\citep{zhang2021roach}. $\lambda_F$ is a tunable loss weight.

For the control prediction branch, we model the action as a beta distribution. The loss $\mathcal{L}_{ctl}$ is:
\begin{align}
\begin{split}
    \mathcal{L}_{ctl} = &  \rm{{\bf KL}\left(Beta({\bf a}_0)||Beta(\hat{{\bf a}}_0)\right)} + \frac{1}{\mathit{K}} \sum_{\mathit{t}=1}^\mathit{K}\rm{{\bf KL}\left(Beta({\bf a}_t)||Beta(\hat{{\bf a}}_t)\right)} \\
    &+ \lambda_F \cdot \mathcal{L}_F\left(\rm{\bf j}^{ctl}_0, \rm{\bf j}^{Expert}_0\right) + \frac{1}{K} \sum_{t=1}^K\mathcal{L}_F\left(\rm{\bf j}^{ctl}_t, \rm{\bf j}^{Expert}_t\right),
\end{split}
\end{align}
where $\rm{Beta({\bf a})}$ denotes the beta distribution represented by the corresponding  predicted distribution parameters and KL-divergence is used to measure the similarity between the predicted control distribution and the one from expert, \textit{i.e.}, $\rm{Beta(\hat{{\bf a}})}$. Feature loss is applied here as well. 
Note that all losses for future time steps ($t \geq 1$) are averaged and then added to the loss for the current time step ($t = 0$), since the action executed immediately should be our key target to optimize.

To help the agent better estimate its current state, we add a speed prediction head to predict current speed $s$ from the image feature and a value prediction head to predict the expected return estimated by the expert, similarly as in \cite{zhang2021roach}. We take the $L_1$ loss for the speed prediction and $L_2$ loss for the value prediction, denoting their weighted sum as $\mathcal{L}_{aux}$.

The overall loss is as follows, as weighted by $\lambda_{traj}, \lambda_{ctl}, \lambda_{aux}$:
\begin{equation}
    \mathcal{L}  = \lambda_{traj} \cdot \mathcal{L}_{traj} + \lambda_{ctl} \cdot \mathcal{L}_{ctl} +  \lambda_{aux} \cdot \mathcal{L}_{aux}.
\end{equation}
    
\subsection{Output Fusion} \label{sec:ensemble}
\begin{wrapfigure}{R}{0.52\textwidth}

\footnotesize
\raisebox{0pt}[\dimexpr\height-2\baselineskip\relax]{
\begin{algorithm}[H]
\SetKwInOut{Parameter}{Hyper parameters}
\SetAlgoLined
\KwIn {sensory input $\rm{{\bf i}}$, speed of the ego vehicle $v$, high level navigation information $\rm{{\bf g}}$.}
\Parameter{combination weight $\alpha \in [0, 0.5]$}
\KwOut {final control signals $\rm{{\bf a}}$}
\BlankLine

$\{\rm{\bf wp}_t\}_{t=0}^K$, $\rm{{\bf a}^{ctl}}$ $\leftarrow$ TCP($\rm{{\bf i}}$, $v$, $\rm{{\bf g}}$)

$\rm{{\bf a}^{traj}}$ $\leftarrow$ Low-level Controller ($\{\rm{\bf wp}_t\}_{t=0}^K$)

Get current $situation$

\eIf{$situation$ {\rm is} trajectory specialized}
{$\rm{{\bf a}}$ $\leftarrow$ $\alpha \times \rm{{\bf a}^{ctl}} + (1-\alpha ) \times \rm{{\bf a}^{traj}}$}
{$\rm{{\bf a}}$ $\leftarrow$ $\alpha \times \rm{{\bf a}^{traj}} + (1- \alpha ) \times \rm{{\bf a}^{ctl}}$}

 \caption{\footnotesize Situation based fusion scheme to combine the two output paradigms}
 \label{alg:ensemble}
\end{algorithm}
}
\end{wrapfigure}

We have two forms of output representations from our \algname framework: the planned trajectory and the predicted control. To further combine their advantages, we devise a situation-based fusion strategy as depicted in Algorithm~\ref{alg:ensemble}. Specifically, denote $\alpha$ as a combination weight whose value is between 0 to 0.5, in a certain situation where one representation is more suitable according to our prior belief, we combine the results from trajectory and control predictions by taking average with weight $\alpha$ so that the more suitable one takes up more weight ($1-\alpha$). Note that the combination weight $\alpha$ indeed does not need to be a constant or symmetric, which means we can set it to different values under different situations or different for specific control signals. In our experiment, we choose the $situation$ according to whether the ego vehicle is turning, implying that if it is turning, the $situation$ is $control \, specialized$ otherwise $trajectory \, specialized$.

\section{Experiments} \label{sec:exp}
\subsection{Experimental Setup} \label{sec:exp_setup}
\textbf{Task \& Evaluation metrics.}
Our method is validated and tested in the CARLA driving simulator \citep{Dosovitskiy17carla}. Given a route defined by a sequence of sparse navigation points together with high level commands (straight, turn left/right, lane changing, and lane following), the closed-loop driving task requires the autonomous agent to drive towards the destination point. It is designed to simulate realistic traffic situations and includes different challenging scenarios such as obstacle avoidance, crossing an unsignalized intersection, and sudden control loss.
%
%
There are three major metrics: Driving Score, Route Completion, and Infraction Score. Route Completion is the percentage of the route completed by the autonomous agent. Infraction Score measures the number of infractions made along the route, with pedestrians, vehicles, road layouts, red lights, and \textit{etc}. Driving Score is the main metric which is the product of Route Completion and Infraction Score.

\textbf{Dataset.}
We use randomly generated routes under random weather conditions to collect 420K data in the 8 public towns offered by the CARLA simulator. Similar to \citep{chen2022lav}, we train \algname on 189K of data in 4 out of 8 towns (Town01, Town03, Town04, and Town06) for ablations and train with all 420K data for our online leaderboard submission.


\subsection{State-of-the-art Comparison} \label{sec:exp_comparison}

\begin{table}[t!]
\centering
\caption{Evaluation on the public CARLA Leaderboard \citep{carlaleaderboard} (accessed in May 2022). Our method \algname and \algname-Ens achieve a driving score of 69.714 and 75.137 respectively with only a monocular camera. More detailed infraction statistics can be found in the Supplementary.}
\label{table:comparesota}
\scalebox{0.85}{
\begin{tabular}{@{}llccccc@{}}
\toprule
\multirow{2}{*}{Rank} & \multirow{2}{*}{Method} & \multicolumn{2}{c}{Sensor Inputs}  & \multicolumn{3}{c}{Key Metrics $\uparrow$}  \\
\cmidrule(r){3-4} \cmidrule(r){5-7}
 & &  \#Cameras & LiDAR & \begin{tabular}[c]{@{}c@{}}Driving \\ Score\end{tabular} & \begin{tabular}[c]{@{}c@{}}Route \\ Completion\end{tabular} & \begin{tabular}[c]{@{}c@{}}Infraction \\ Score\end{tabular} \\ \midrule
1    & \textbf{\algname-Ens} (ours)  & 1  &\xmark &\textbf{75.137} & 85.629 & \textbf{0.873} \\
1    & \textbf{\algname} (ours)      & 1  &\xmark & \textbf{69.714} & 82.962 & \textbf{0.851} \\
1    & \textbf{\algname-SB} (ours)      & 1  &\xmark & \textbf{68.695} & 82.957 & \textbf{0.833} \\ \midrule
2    & LAV \citep{chen2022lav}  & 4  &\cmark & 61.846 & \textbf{94.459} & 0.640 \\
3    & Transfuser        & 3  &\cmark & 61.181 & 86.694          & 0.714 \\
4    & Latent Transfuser & 3  &\xmark & 45.029 & 75.366          & 0.618 \\
5    & GRIAD \citep{chekroun2021gri} & 3  &\xmark & 36.787 & 61.855          & 0.597 \\
6    & Transfuser+ \citep{jaeger2021transfuser+}      & 4  &\cmark & 34.577 & 69.841          & 0.562 \\
7    & WoR \citep{chen2021wor}               & 4  &\xmark & 31.370 & 57.647          & 0.557 \\
8    & MaRLn \citep{toromanoff2020modelfreerl}             & 1  &\xmark & 24.980 & 46.968          & 0.518 \\ 
9    & NEAT \citep{chitta2021neat}   & 3 & \xmark & 21.832 & 41.707 & 0.650 \\ \bottomrule
\end{tabular}
}
\end{table}

Table~\ref{table:comparesota} shows the result of the comparison between our method and the top 8 entries on the public CARLA Leaderboard \citep{carlaleaderboard}. We report the results of \algname and two variants. \textbf{\algname-SB} replaces shared encoders of \algname with two separate ones for two branches, and \textbf{\algname-Ens} is the ensemble of \algname and \algname-SB.
Our method \algname-Ens ranks first on the leaderboard with a 75.137 driving score and highest infraction score, and \algname alone also surpasses prior methods.
Note that our method only uses a monocular camera while the top 2-4 methods all use multiple cameras and a LiDAR. Our driving score is 50.157 higher than the second-best monocular camera method, MaRLn \citep{toromanoff2020modelfreerl}.
Our route completion is slightly inferior to the LiDAR candidates - one reason is that methods using LiDAR may have a better object detection ability. Based on the detection results, they usually adopt a crawling strategy, indicating that the vehicle would move slowly when it has stopped for a long time and there are no obstacles ahead. As described in \citep{jaeger2021transfuser+}, this could alleviate ego vehicle's blocking problems to boost the route completion performance.

\subsection{Control vs. Trajectory} \label{sec:ctl_vs_traj}
\begin{table}[t!]
\centering
\caption{Comparison between the control and trajectory only model in terms of infractions frequency. TurnRatio means the corresponding ratio of happening during turning.}
\label{table:trajvscontrol}
\scalebox{0.8}{
\begin{tabular}{@{}lccccccccc@{}}
\toprule
\multirow{2}{*}{Model} & \multirow{2}{*}{\begin{tabular}[c]{@{}c@{}}Driving \\ Score\end{tabular}}
& \multicolumn{2}{c}{Collisions vehicles}                                                      & \multicolumn{2}{c}{Collisions layout}                                                        & \multicolumn{2}{c}{Off-road infractions}                                                     & \multicolumn{2}{c}{Agent blocked}                                                            \\ \cmidrule(l){3-4} \cmidrule(l){5-6} \cmidrule(l){7-8} \cmidrule(l){9-10} 
& & \#/km  $\downarrow$ & TurnRatio & \#/km  $\downarrow$ & TurnRatio& \#/km  $\downarrow$ & TurnRatio& \#/km  $\downarrow$ & TurnRatio \\ \midrule
Control-Only   & 32.45$\pm$2.23 & 1.25                       & 50.90\%                                                        & \textbf{0.23}                       & 10.00\%                                                           & \textbf{0.59}                       & 46.15\%                                                       & \textbf{0.41}                       & 50.00\%                                                           \\
Trajectory-Only & 28.29$\pm$3.03 & \textbf{0.85}                      & 38.70\%                                                        & 0.77                       & 64.20\%                                                        & 0.74                       & 62.90\%                                                        & 0.77                        & 64.20\%                                                        \\ \bottomrule
\end{tabular}
}
\end{table}

In this section, we conduct quantitative experiments to compare the \textbf{Control-Only} model and the \textbf{Trajectory-Only} model to demonstrate their advantages and disadvantages.
For both models, we use the same setting except for the output head and its corresponding loss. We use a ResNet-34 to encode visual inputs and a measurement module to encode the navigation information. Similar to~\citep{zhang2021roach}, we add speed and value heads as auxiliary tasks to help the model better encode the environment.
For \textbf{Control-Only}, we predict the control distribution based on the concatenated latent feature from the two encoders.
%
%
As for \textbf{Trajectory-Only}, we feed the feature to a GRU decoder to generate waypoints. 
%
As shown in Table~\ref{table:trajvscontrol}, though Trajectory-Only collides with vehicles less frequently than  Control-Only, it has more layout collisions, off-road infractions, and agent blocks.
We also count the ratio of each kind of infraction that occurs during turning. It can be observed that for Trajectory-Only, a large portion of such infractions happen when the ego agent is turning compared to Control-Only.
This has verified that Trajectory-Only performs worse when the agent is turning, which is probably caused by the unsatisfactory trajectory following performance of simple PID controllers as discussed in Sec.~\ref{sec:intro}.
As for the fact that Control-Only has a higher vehicle-collision rate, it is because the model focuses on the current time step and the reaction to potential collisions tends to be late, as depicted in Sec.~\ref{sec:intro} as well. The results above further validate the necessity of combining the two output paradigms.

\subsection{Ablative Study and Visualization} \label{sec:ablation}

\textbf{Component analysis.} We first validate the effectiveness of the trajectory-guided multi-step control prediction design, as shown in Table~\ref{table:ablation}.
We only employ the control branch output except for the last complete one when fusion is applied for these ablations. Adding a trajectory branch as an auxiliary task improves the performance by 2.5 points. The multi-step predictions with our temporal module greatly help with 7.9 points gain, and adding the trajectory-guided attention further acquires an improvement of 3.2 points. Finally, applying our situation based fusion scheme ($\alpha$ is set to 0.3) significantly boosts the infraction score, leading the overall driving score to 57.

\begin{figure}[t!]
    \centering
    \includegraphics[width=1.0\textwidth]{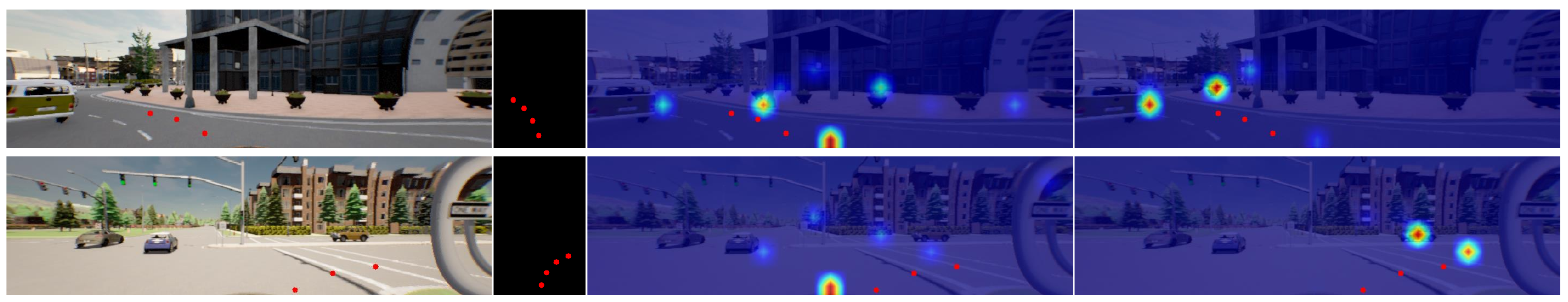}
    \caption{The trajectory-guided attention maps in two cases. In each case (row), from the left to right we show that the input image with the predicted trajectory (the first waypoint is projected out of the image), the predicted trajectory in the top-down view, the attention map $\rm{\bf w}_1$, the attention map $\rm{\bf w}_3$.}
    \label{fig:att}
\end{figure}
\begin{table}[t!]
\parbox[b]{.45\linewidth}{
\centering
\caption{Ablative study on the effectiveness of different components design of our model.}
\label{table:ablation}
\scalebox{0.85}{
\begin{tabular}[b]{@{}lccc@{}}
\toprule
Exp. & \multicolumn{1}{c}{\begin{tabular}[c]{@{}c@{}}Driving \\ Score\end{tabular}} & \multicolumn{1}{c}{\begin{tabular}[c]{@{}c@{}}Route \\ Completion\end{tabular}} & \multicolumn{1}{c}{\begin{tabular}[c]{@{}c@{}}Infraction \\ Score\end{tabular}} \\ \midrule
Control   & 32.45$\pm$2.23 & 76.54$\pm$3.22 & 0.45$\pm$0.03  \\
$+$ traj-task & 34.98$\pm$1.96 & 81.32$\pm$5.50 & 0.49$\pm$0.05 \\
$+$ temporal & 42.87$\pm$4.77  & \textbf{87.51}$\pm$3.63 & 0.49$\pm$0.07 \\
$+$ traj-attn & 46.08$\pm$3.47 & 84.95$\pm$1.84 & 0.56$\pm$0.03  \\
$+$ fusion  & \textbf{57.01}$\pm$1.88 & 85.27$\pm$1.20  & \textbf{0.67}$\pm$0.01    \\ 
\bottomrule
\end{tabular}}
}
\hspace{15pt}
\parbox[b]{.45\linewidth}{
\centering
\caption{Comparison between MTL and ensemble methods ($\alpha$ is 0.3 for all experiments).}
\label{table:mtlvsensemble}
\scalebox{0.85}{
\begin{tabular}[b]{@{}lcccc@{}}
\toprule
Exp.                                                                             & \begin{tabular}[c]{@{}c@{}}Driving \\ Score\end{tabular}  & \#Param. & FLOPs & FPS \\ \midrule
Ensemble  &45.03$\pm$1.28 & 46.81M & 17.07G & 69.47\\
MTL & 48.27$\pm$0.58 & 23.58M & 8.54G & 133.30        \\ 
\algname-SB  &52.46$\pm$4.66 & 47.26M & 17.07G & 69.35  \\
\algname & 57.01$\pm$1.88& 25.77M & 8.54G & 125.71 \\ 
\algname-Ens &59.09$\pm$3.66 & 73.03M & 25.61G & 44.70  \\
\bottomrule
\end{tabular}}}
\end{table}

\textbf{Multi-task vs. Ensemble.} 
The comparison regarding their performances and computational complexity is given in Table~\ref{table:mtlvsensemble}.
\textbf{Ensemble} denotes directly combining the outputs of Control-Only and Trajectory-Only with our situation based fusion scheme.
\textbf{MTL} represents the model with a shared CNN backbone and measurement encoders followed by a trajectory branch and a control branch, but the control branch predicts current step prediction only and there are no interactions between the two branches.
%
%
We conclude that directly combining two models with our fusion scheme greatly improves the performance, and using an MTL approach works better than ensemble but with a much smaller model size and GFLOPs. A conventional ensemble approach to combine results from \algname and \algname-SB as \algname-Ens brings further performance gain at the cost of computational complexity.

\captionsetup[wrapfigure]{font=footnotesize}
\begin{wrapfigure}{r}{0pt}
        \includegraphics[width=0.38\textwidth]{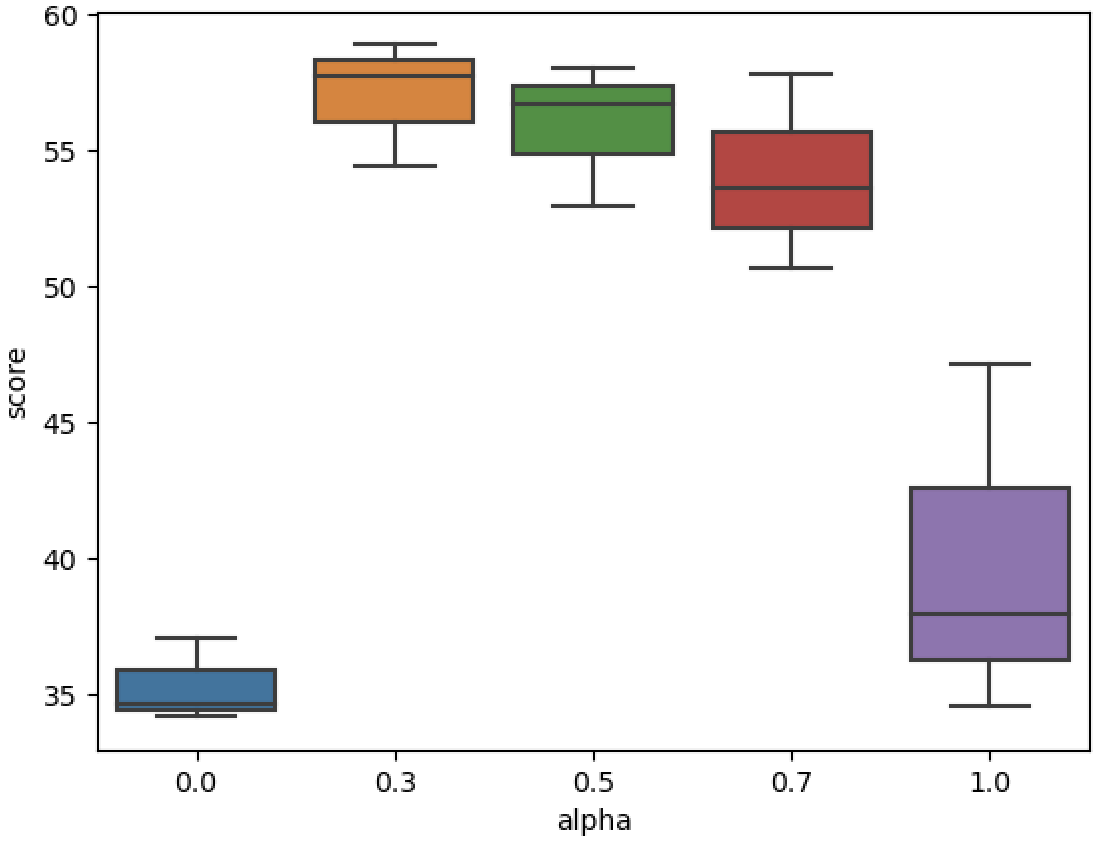}
    \caption{Box plot of the driving score with different $\alpha$ values (3 trials for each $\alpha$).}
    \label{fig:mixture_weight}
\end{wrapfigure}
\textbf{Situation based fusion weight.}
We investigate the choice of the combination weight $\alpha$ in the situation based fusion scheme and show the box plot of the driving scores in the figure to the right. Besides $\alpha \in [0,0.5]$, we additionally test 0.7 and 1, meaning that two results are conversely mixed with our $specialization$ definition.
We see that only using the control from the specialized branch $\alpha = 0$ performs poorly while directly taking the average or fusing conversely still has comparable results. One reason is that the $situation$ criterion used here is whether the vehicle is turning, making most cases $trajectory\ specialized$, and the stronger control branch is not utilized enough if $\alpha$ is small.
Note that the situation based fusion scheme is general and flexible, and the criterion or $\alpha$ value used here is relatively coarse. 

\textbf{Visualization.} Fig.~\ref{fig:att} visualizes the trajectory-guided attention maps. The trajectory branch provides location-related information to guide the control branch to focus on important regions which are useful for future control prediction. See more qualitative results in the Supplementary.

\section{Conclusion}
In this work, we study two learning and prediction paradigms based on trajectory and direct control, respectively, for end-to-end autonomous driving. We propose a unified framework comprised of a trajectory branch and a novel multi-step control branch with interactions in between. We design a situation based fusion scheme to combine the results from two branches. Our method with only a monocular camera has achieved state-of-the-art performance on the CARLA Leaderboard.

\section*{Acknowledgments}
This work was partly supported by National Key Research and Development Program of China (2020AAA0107600), NSFC (62206172, 61972250), Shanghai Municipal Science and Technology Major Project (2021SHZDZX0102), and Shanghai Committee of Science and Technology (21DZ1100100, 22511105100).

\bibliographystyle{plainnat}
{
\small
\bibliography{egbib}

\begin{thebibliography}{63}
\providecommand{\natexlab}[1]{#1}
\providecommand{\url}[1]{\texttt{#1}}
\expandafter\ifx\csname urlstyle\endcsname\relax
  \providecommand{\doi}[1]{doi: #1}\else
  \providecommand{\doi}{doi: \begingroup \urlstyle{rm}\Url}\fi

\bibitem[car(2022)]{carlaleaderboard}
{CARLA} autonomous driving leaderboard.
\newblock \url{https://leaderboard.carla.org/}, 2022.

\bibitem[Argyriou et~al.(2006)Argyriou, Evgeniou, and Pontil]{argyriou2006mtfl}
Andreas Argyriou, Theodoros Evgeniou, and Massimiliano Pontil.
\newblock Multi-task feature learning.
\newblock In \emph{NeurIPS}, 2006.

\bibitem[Bojarski et~al.(2016)Bojarski, Del~Testa, Dworakowski, Firner, Flepp,
  Goyal, Jackel, Monfort, Muller, Zhang, et~al.]{bojarski2016nvend}
Mariusz Bojarski, Davide Del~Testa, Daniel Dworakowski, Bernhard Firner, Beat
  Flepp, Prasoon Goyal, Lawrence~D Jackel, Mathew Monfort, Urs Muller, Jiakai
  Zhang, et~al.
\newblock End to end learning for self-driving cars.
\newblock \emph{arXiv preprint arXiv:1604.07316}, 2016.

\bibitem[Bojarski et~al.(2020)Bojarski, Chen, Daw, De{\u{g}}irmenci, Deri,
  Firner, Flepp, Gogri, Hong, Jackel, et~al.]{pilotnet2020nvidia}
Mariusz Bojarski, Chenyi Chen, Joyjit Daw, Alperen De{\u{g}}irmenci, Joya Deri,
  Bernhard Firner, Beat Flepp, Sachin Gogri, Jesse Hong, Lawrence Jackel,
  et~al.
\newblock The nvidia pilotnet experiments.
\newblock \emph{arXiv preprint arXiv:2010.08776}, 2020.

\bibitem[Camacho and Alba(2013)]{camacho2013mpc}
Eduardo~F Camacho and Carlos~Bordons Alba.
\newblock \emph{Model predictive control}.
\newblock Springer science \& business media, 2013.

\bibitem[Carranza-Garc{\'\i}a et~al.(2021)Carranza-Garc{\'\i}a,
  Lara-Ben{\'\i}tez, Garc{\'\i}a-Guti{\'e}rrez, and
  Riquelme]{carranza2021odadensemble}
Manuel Carranza-Garc{\'\i}a, Pedro Lara-Ben{\'\i}tez, Jorge
  Garc{\'\i}a-Guti{\'e}rrez, and Jos{\'e}~C Riquelme.
\newblock Enhancing object detection for autonomous driving by optimizing
  anchor generation and addressing class imbalance.
\newblock \emph{Neurocomputing}, 2021.

\bibitem[Caruana(1997)]{caruana1997mtl}
Rich Caruana.
\newblock Multitask learning.
\newblock \emph{Machine learning}, 1997.

\bibitem[Casas et~al.(2021)Casas, Sadat, and Urtasun]{casas2021mp3}
Sergio Casas, Abbas Sadat, and Raquel Urtasun.
\newblock Mp3: A unified model to map, perceive, predict and plan.
\newblock In \emph{CVPR}, 2021.

\bibitem[Chekroun et~al.(2021)Chekroun, Toromanoff, Hornauer, and
  Moutarde]{chekroun2021gri}
Raphael Chekroun, Marin Toromanoff, Sascha Hornauer, and Fabien Moutarde.
\newblock Gri: General reinforced imitation and its application to vision-based
  autonomous driving.
\newblock \emph{arXiv preprint arXiv:2111.08575}, 2021.

\bibitem[Chen and Kr{\"a}henb{\"u}hl(2022)]{chen2022lav}
Dian Chen and Philipp Kr{\"a}henb{\"u}hl.
\newblock Learning from all vehicles.
\newblock In \emph{CVPR}, 2022.

\bibitem[Chen et~al.(2020)Chen, Zhou, Koltun, and
  Kr{\"a}henb{\"u}hl]{chen2020lbc}
Dian Chen, Brady Zhou, Vladlen Koltun, and Philipp Kr{\"a}henb{\"u}hl.
\newblock Learning by cheating.
\newblock In \emph{CoRL}, 2020.

\bibitem[Chen et~al.(2021)Chen, Koltun, and Kr{\"a}henb{\"u}hl]{chen2021wor}
Dian Chen, Vladlen Koltun, and Philipp Kr{\"a}henb{\"u}hl.
\newblock Learning to drive from a world on rails.
\newblock In \emph{ICCV}, 2021.

\bibitem[Chen et~al.(2022)Chen, Sima, Li, Zheng, Xu, Geng, Li, He, Shi, Qiao,
  et~al.]{chen2022persformer}
Li~Chen, Chonghao Sima, Yang Li, Zehan Zheng, Jiajie Xu, Xiangwei Geng,
  Hongyang Li, Conghui He, Jianping Shi, Yu~Qiao, et~al.
\newblock Persformer: 3d lane detection via perspective transformer and the
  openlane benchmark.
\newblock \emph{arXiv preprint arXiv:2203.11089}, 2022.

\bibitem[Chennupati et~al.(2019)Chennupati, Sistu, Yogamani, and
  A~Rawashdeh]{chennupati2019multinet++}
Sumanth Chennupati, Ganesh Sistu, Senthil Yogamani, and Samir A~Rawashdeh.
\newblock Multinet++: Multi-stream feature aggregation and geometric loss
  strategy for multi-task learning.
\newblock In \emph{CVPRW}, 2019.

\bibitem[Chitta et~al.(2021)Chitta, Prakash, and Geiger]{chitta2021neat}
Kashyap Chitta, Aditya Prakash, and Andreas Geiger.
\newblock Neat: Neural attention fields for end-to-end autonomous driving.
\newblock In \emph{ICCV}, 2021.

\bibitem[Cho et~al.(2014)Cho, van Merrienboer, Çaglar G{\"u}lçehre, Bahdanau,
  Bougares, Schwenk, and Bengio]{GRU}
Kyunghyun Cho, Bart van Merrienboer, Çaglar G{\"u}lçehre, Dzmitry Bahdanau,
  Fethi Bougares, Holger Schwenk, and Yoshua Bengio.
\newblock Learning phrase representations using rnn encoder–decoder for
  statistical machine translation.
\newblock In \emph{EMNLP}, 2014.

\bibitem[Codevilla et~al.(2018)Codevilla, M{\"u}ller, L{\'o}pez, Koltun, and
  Dosovitskiy]{codevilla2018cil}
Felipe Codevilla, Matthias M{\"u}ller, Antonio L{\'o}pez, Vladlen Koltun, and
  Alexey Dosovitskiy.
\newblock End-to-end driving via conditional imitation learning.
\newblock In \emph{ICRA}, 2018.

\bibitem[Codevilla et~al.(2019)Codevilla, Santana, L{\'o}pez, and
  Gaidon]{codevilla2019cilrs}
Felipe Codevilla, Eder Santana, Antonio~M L{\'o}pez, and Adrien Gaidon.
\newblock Exploring the limitations of behavior cloning for autonomous driving.
\newblock In \emph{ICCV}, 2019.

\bibitem[Dietterich(2000)]{dietterich2000ensemble}
Thomas~G Dietterich.
\newblock Ensemble methods in machine learning.
\newblock In \emph{MCS}, 2000.

\bibitem[Dosovitskiy et~al.(2017)Dosovitskiy, Ros, Codevilla, Lopez, and
  Koltun]{Dosovitskiy17carla}
Alexey Dosovitskiy, German Ros, Felipe Codevilla, Antonio Lopez, and Vladlen
  Koltun.
\newblock {CARLA}: {An} open urban driving simulator.
\newblock In \emph{CoRL}, 2017.

\bibitem[Guo et~al.(2019)Guo, Cao, Chen, Sun, and Hu]{guo2019mpcfollowing}
Hongyan Guo, Dongpu Cao, Hong Chen, Zhenping Sun, and Yunfeng Hu.
\newblock Model predictive path following control for autonomous cars
  considering a measurable disturbance: Implementation, testing, and
  verification.
\newblock \emph{MSSP}, 2019.

\bibitem[Hawke et~al.(2020)Hawke, Shen, Gurau, Sharma, Reda, Nikolov, Mazur,
  Micklethwaite, Griffiths, Shah, et~al.]{hawke2020urbancil}
Jeffrey Hawke, Richard Shen, Corina Gurau, Siddharth Sharma, Daniele Reda,
  Nikolay Nikolov, Przemys{\l}aw Mazur, Sean Micklethwaite, Nicolas Griffiths,
  Amar Shah, et~al.
\newblock Urban driving with conditional imitation learning.
\newblock In \emph{ICRA}, 2020.

\bibitem[He et~al.(2016)He, Zhang, Ren, and Sun]{he2016resnet}
Kaiming He, Xiangyu Zhang, Shaoqing Ren, and Jian Sun.
\newblock Deep residual learning for image recognition.
\newblock In \emph{CVPR}, 2016.

\bibitem[Hou et~al.(2019)Hou, Ma, Liu, and Loy]{hou2019fmnet}
Yuenan Hou, Zheng Ma, Chunxiao Liu, and Chen~Change Loy.
\newblock Learning to steer by mimicking features from heterogeneous auxiliary
  networks.
\newblock In \emph{AAAI}, 2019.

\bibitem[Hu et~al.(2022)Hu, Chen, Wu, Li, Yan, and Tao]{hu2022stp3}
Shengchao Hu, Li~Chen, Penghao Wu, Hongyang Li, Junchi Yan, and Dacheng Tao.
\newblock St-p3: End-to-end vision-based autonomous driving via
  spatial-temporal feature learning.
\newblock In \emph{ECCV}, 2022.

\bibitem[Huch et~al.(2021)Huch, Ongel, Betz, and Lienkamp]{huch2021e2ecav}
Sebastian Huch, Aybike Ongel, Johannes Betz, and Markus Lienkamp.
\newblock Multi-task end-to-end self-driving architecture for cav platoons.
\newblock \emph{Sensors}, 2021.

\bibitem[Ishihara et~al.(2021)Ishihara, Kanervisto, Miura, and
  Hautamaki]{ishihara2021multie2eattention}
Keishi Ishihara, Anssi Kanervisto, Jun Miura, and Ville Hautamaki.
\newblock Multi-task learning with attention for end-to-end autonomous driving.
\newblock In \emph{CVPR}, 2021.

\bibitem[Jacobs et~al.(1991)Jacobs, Jordan, Nowlan, and
  Hinton]{jacobs1991expertmixture}
Robert~A Jacobs, Michael~I Jordan, Steven~J Nowlan, and Geoffrey~E Hinton.
\newblock Adaptive mixtures of local experts.
\newblock \emph{Neural computation}, 1991.

\bibitem[Jaeger(2021)]{jaeger2021transfuser+}
Bernhard Jaeger.
\newblock Expert drivers for autonomous driving.
\newblock Master's thesis, University of Tübingen, 2021.

\bibitem[Kendall et~al.(2019)Kendall, Hawke, Janz, Mazur, Reda, Allen, Lam,
  Bewley, and Shah]{kendall2019driveaday}
Alex Kendall, Jeffrey Hawke, David Janz, Przemyslaw Mazur, Daniele Reda,
  John-Mark Allen, Vinh-Dieu Lam, Alex Bewley, and Amar Shah.
\newblock Learning to drive in a day.
\newblock In \emph{ICRA}, 2019.

\bibitem[Kim et~al.(2020)Kim, Lee, Lee, Lee, and Kim]{kim2020fasnet}
Inhan Kim, Hyemin Lee, Joonyeong Lee, Eunseop Lee, and Daijin Kim.
\newblock Multi-task learning with future states for vision-based autonomous
  driving.
\newblock In \emph{ACCV}, 2020.

\bibitem[Kim et~al.(2022)Kim, Lee, and Kim]{kim2022mode}
Inhan Kim, Joonyeong Lee, and Daijin Kim.
\newblock Learning mixture of domain-specific experts via disentangled factors
  for autonomous driving.
\newblock In \emph{AAAI}, 2022.

\bibitem[Kr{\"a}henb{\"u}hl and Koltun(2015)]{krahenbuhl2015lpo}
Philipp Kr{\"a}henb{\"u}hl and Vladlen Koltun.
\newblock Learning to propose objects.
\newblock In \emph{CVPR}, 2015.

\bibitem[Kumar et~al.(2021)Kumar, Yogamani, Rashed, Sitsu, Witt, Leang, Milz,
  and M{\"a}der]{kumar2021omnidet}
Varun~Ravi Kumar, Senthil Yogamani, Hazem Rashed, Ganesh Sitsu, Christian Witt,
  Isabelle Leang, Stefan Milz, and Patrick M{\"a}der.
\newblock Omnidet: Surround view cameras based multi-task visual perception
  network for autonomous driving.
\newblock \emph{RA-L}, 2021.

\bibitem[Lakshminarayanan et~al.(2017)Lakshminarayanan, Pritzel, and
  Blundell]{lakshminarayanan2017deepensemble}
Balaji Lakshminarayanan, Alexander Pritzel, and Charles Blundell.
\newblock Simple and scalable predictive uncertainty estimation using deep
  ensembles.
\newblock In \emph{NeurIPS}, 2017.

\bibitem[Li et~al.(2022)Li, Yu, Meng, Caine, Ngiam, Peng, Shen, Wu, Lu, Zhou,
  et~al.]{li2022deepfusion}
Yingwei Li, Adams~Wei Yu, Tianjian Meng, Ben Caine, Jiquan Ngiam, Daiyi Peng,
  Junyang Shen, Bo~Wu, Yifeng Lu, Denny Zhou, et~al.
\newblock Deepfusion: Lidar-camera deep fusion for multi-modal 3d object
  detection.
\newblock In \emph{CVPR}, 2022.

\bibitem[Li et~al.(2018)Li, Motoyoshi, Sasaki, Ogata, and
  Sugano]{li2018rethinking}
Zhihao Li, Toshiyuki Motoyoshi, Kazuma Sasaki, Tetsuya Ogata, and Shigeki
  Sugano.
\newblock Rethinking self-driving: Multi-task knowledge for better
  generalization and accident explanation ability.
\newblock \emph{arXiv preprint arXiv:1809.11100}, 2018.

\bibitem[Liang et~al.(2019)Liang, Yang, Chen, Hu, and
  Urtasun]{liang2019multi3d}
Ming Liang, Bin Yang, Yun Chen, Rui Hu, and Raquel Urtasun.
\newblock Multi-task multi-sensor fusion for 3d object detection.
\newblock In \emph{CVPR}, 2019.

\bibitem[Liang et~al.(2018)Liang, Wang, Yang, and Xing]{liang2018cirl}
Xiaodan Liang, Tairui Wang, Luona Yang, and Eric Xing.
\newblock Cirl: Controllable imitative reinforcement learning for vision-based
  self-driving.
\newblock In \emph{ECCV}, 2018.

\bibitem[Muller et~al.(2005{\natexlab{a}})Muller, Ben, Cosatto, Flepp, and
  Cun]{muller2005e2eavoid}
Urs Muller, Jan Ben, Eric Cosatto, Beat Flepp, and Yann Cun.
\newblock Off-road obstacle avoidance through end-to-end learning.
\newblock In \emph{NeurIPS}, 2005{\natexlab{a}}.

\bibitem[Muller et~al.(2005{\natexlab{b}})Muller, Ben, Cosatto, Flepp, and
  Cun]{muller2005e2eavoidance}
Urs Muller, Jan Ben, Eric Cosatto, Beat Flepp, and Yann Cun.
\newblock Off-road obstacle avoidance through end-to-end learning.
\newblock In \emph{NeurIPS}, 2005{\natexlab{b}}.

\bibitem[Ohn-Bar et~al.(2020)Ohn-Bar, Prakash, Behl, Chitta, and
  Geiger]{ohn2020lsd}
Eshed Ohn-Bar, Aditya Prakash, Aseem Behl, Kashyap Chitta, and Andreas Geiger.
\newblock Learning situational driving.
\newblock In \emph{CVPR}, 2020.

\bibitem[Pokle et~al.(2019)Pokle, Mart{\'\i}n-Mart{\'\i}n, Goebel, Chow, Ewald,
  Yang, Wang, Sadeghian, Sadigh, Savarese, et~al.]{pokle2019robot}
Ashwini Pokle, Roberto Mart{\'\i}n-Mart{\'\i}n, Patrick Goebel, Vincent Chow,
  Hans~M Ewald, Junwei Yang, Zhenkai Wang, Amir Sadeghian, Dorsa Sadigh, Silvio
  Savarese, et~al.
\newblock Deep local trajectory replanning and control for robot navigation.
\newblock In \emph{ICRA}, 2019.

\bibitem[Pomerleau(1988)]{pomerleau1988alvinn}
Dean~A Pomerleau.
\newblock Alvinn: An autonomous land vehicle in a neural network.
\newblock In \emph{NeurIPS}, 1988.

\bibitem[Prakash et~al.(2020)Prakash, Behl, Ohn-Bar, Chitta, and
  Geiger]{prakash2020darb}
Aditya Prakash, Aseem Behl, Eshed Ohn-Bar, Kashyap Chitta, and Andreas Geiger.
\newblock Exploring data aggregation in policy learning for vision-based urban
  autonomous driving.
\newblock In \emph{CVPR}, 2020.

\bibitem[Prakash et~al.(2021)Prakash, Chitta, and
  Geiger]{prakash2021transfuser}
Aditya Prakash, Kashyap Chitta, and Andreas Geiger.
\newblock Multi-modal fusion transformer for end-to-end autonomous driving.
\newblock In \emph{CVPR}, 2021.

\bibitem[Rajamanoharan et~al.(2019)Rajamanoharan, Kanac{\i}, Li, Gong,
  et~al.]{rajamanoharan2019mtmlreid}
Georgia Rajamanoharan, Ayta{\c{c}} Kanac{\i}, Minxian Li, Shaogang Gong, et~al.
\newblock Multi-task mutual learning for vehicle re-identification.
\newblock In \emph{CVPR}, 2019.

\bibitem[Rhinehart et~al.(2019)Rhinehart, McAllister, and
  Levine]{rhinehart2019deep}
Nicholas Rhinehart, Rowan McAllister, and Sergey Levine.
\newblock Deep imitative models for flexible inference, planning, and control.
\newblock In \emph{ICLR}, 2019.

\bibitem[Sadat et~al.(2020)Sadat, Casas, Ren, Wu, Dhawan, and
  Urtasun]{sadat2020p3}
Abbas Sadat, Sergio Casas, Mengye Ren, Xinyu Wu, Pranaab Dhawan, and Raquel
  Urtasun.
\newblock Perceive, predict, and plan: Safe motion planning through
  interpretable semantic representations.
\newblock In \emph{ECCV}, 2020.

\bibitem[Toromanoff et~al.(2020)Toromanoff, Wirbel, and
  Moutarde]{toromanoff2020modelfreerl}
Marin Toromanoff, Emilie Wirbel, and Fabien Moutarde.
\newblock End-to-end model-free reinforcement learning for urban driving using
  implicit affordances.
\newblock In \emph{CVPR}, 2020.

\bibitem[Vyas et~al.(2018)Vyas, Jammalamadaka, Zhu, Das, Kaul, and
  Willke]{vyas2018oodensemble}
Apoorv Vyas, Nataraj Jammalamadaka, Xia Zhu, Dipankar Das, Bharat Kaul, and
  Theodore~L Willke.
\newblock Out-of-distribution detection using an ensemble of self supervised
  leave-out classifiers.
\newblock In \emph{ECCV}, pages 550--564, 2018.

\bibitem[Wu et~al.(2021)Wu, Liao, Zhang, and Wang]{wu2021yolop}
Dong Wu, Manwen Liao, Weitian Zhang, and Xinggang Wang.
\newblock Yolop: You only look once for panoptic driving perception.
\newblock \emph{arXiv preprint arXiv:2108.11250}, 2021.

\bibitem[Xiao et~al.(2020)Xiao, Codevilla, Gurram, Urfalioglu, and
  L{\'o}pez]{xiao2020multimodal}
Yi~Xiao, Felipe Codevilla, Akhil Gurram, Onay Urfalioglu, and Antonio~M
  L{\'o}pez.
\newblock Multimodal end-to-end autonomous driving.
\newblock \emph{T-ITS}, 2020.

\bibitem[Xu et~al.(2017)Xu, Gao, Yu, and Darrell]{xu2017lstme2e}
Huazhe Xu, Yang Gao, Fisher Yu, and Trevor Darrell.
\newblock End-to-end learning of driving models from large-scale video
  datasets.
\newblock In \emph{CVPR}, 2017.

\bibitem[Xu et~al.(2020)Xu, Wang, Wang, and Guo]{xu2020mmensemble}
Jie Xu, Wei Wang, Hanyuan Wang, and Jinhong Guo.
\newblock Multi-model ensemble with rich spatial information for object
  detection.
\newblock \emph{Pattern Recognition}, 2020.

\bibitem[Yang et~al.(2018)Yang, Zhang, Yu, Cai, and Luo]{yang2018e2emmmt}
Zhengyuan Yang, Yixuan Zhang, Jerry Yu, Junjie Cai, and Jiebo Luo.
\newblock End-to-end multi-modal multi-task vehicle control for self-driving
  cars with visual perceptions.
\newblock In \emph{ICPR}, 2018.

\bibitem[Zablocki et~al.(2021)Zablocki, Ben-Younes, P{\'e}rez, and
  Cord]{zablocki2021xai}
{\'E}loi Zablocki, H{\'e}di Ben-Younes, Patrick P{\'e}rez, and Matthieu Cord.
\newblock Explainability of vision-based autonomous driving systems: Review and
  challenges.
\newblock \emph{arXiv preprint arXiv:2101.05307}, 2021.

\bibitem[Zeng et~al.(2019)Zeng, Luo, Suo, Sadat, Yang, Casas, and
  Urtasun]{zeng2019nmp}
Wenyuan Zeng, Wenjie Luo, Simon Suo, Abbas Sadat, Bin Yang, Sergio Casas, and
  Raquel Urtasun.
\newblock End-to-end interpretable neural motion planner.
\newblock In \emph{CVPR}, 2019.

\bibitem[Zhang and Ohn-Bar(2021)]{zhang2021lbw}
Jimuyang Zhang and Eshed Ohn-Bar.
\newblock Learning by watching.
\newblock In \emph{CVPR}, 2021.

\bibitem[Zhang et~al.(2021)Zhang, Liniger, Dai, Yu, and
  Van~Gool]{zhang2021roach}
Zhejun Zhang, Alexander Liniger, Dengxin Dai, Fisher Yu, and Luc Van~Gool.
\newblock End-to-end urban driving by imitating a reinforcement learning coach.
\newblock In \emph{ICCV}, 2021.

\bibitem[Zhao et~al.(2021)Zhao, He, Liang, Huang, Van~den Broeck, and
  Soatto]{zhao2021sam}
Albert Zhao, Tong He, Yitao Liang, Haibin Huang, Guy Van~den Broeck, and
  Stefano Soatto.
\newblock Sam: Squeeze-and-mimic networks for conditional visual driving policy
  learning.
\newblock In \emph{CoRL}, 2021.

\bibitem[Zhao et~al.(2022)Zhao, Wu, Xu, Che, Lu, Tang, and Liu]{zhao2022cadre}
Yinuo Zhao, Kun Wu, Zhiyuan Xu, Zhengping Che, Qi~Lu, Jian Tang, and Chi~Harold
  Liu.
\newblock Cadre: A cascade deep reinforcement learning framework for
  vision-based autonomous urban driving.
\newblock In \emph{AAAI}, 2022.

\bibitem[Zhu and Zhao(2022)]{zhu2022mtcil}
Zeyu Zhu and Huijing Zhao.
\newblock Multi-task conditional imitation learning for autonomous navigation
  at crowded intersections.
\newblock \emph{arXiv preprint arXiv:2202.10124}, 2022.

\end{thebibliography}


\begin{thebibliography}{16}
\providecommand{\natexlab}[1]{#1}
\providecommand{\url}[1]{\texttt{#1}}
\expandafter\ifx\csname urlstyle\endcsname\relax
  \providecommand{\doi}[1]{doi: #1}\else
  \providecommand{\doi}{doi: \begingroup \urlstyle{rm}\Url}\fi

\bibitem[car(2022)]{carlaleaderboard}
{CARLA} autonomous driving leaderboard.
\newblock \url{https://leaderboard.carla.org/}, 2022.

\bibitem[Chekroun et~al.(2021)Chekroun, Toromanoff, Hornauer, and
  Moutarde]{chekroun2021gri}
Raphael Chekroun, Marin Toromanoff, Sascha Hornauer, and Fabien Moutarde.
\newblock Gri: General reinforced imitation and its application to vision-based
  autonomous driving.
\newblock \emph{arXiv preprint arXiv:2111.08575}, 2021.

\bibitem[Chen and Kr{\"a}henb{\"u}hl(2022)]{chen2022lav}
Dian Chen and Philipp Kr{\"a}henb{\"u}hl.
\newblock Learning from all vehicles.
\newblock In \emph{CVPR}, 2022.

\bibitem[Chen et~al.(2021)Chen, Koltun, and Kr{\"a}henb{\"u}hl]{chen2021wor}
Dian Chen, Vladlen Koltun, and Philipp Kr{\"a}henb{\"u}hl.
\newblock Learning to drive from a world on rails.
\newblock In \emph{ICCV}, 2021.

\bibitem[Chitta et~al.(2021)Chitta, Prakash, and Geiger]{chitta2021neat}
Kashyap Chitta, Aditya Prakash, and Andreas Geiger.
\newblock Neat: Neural attention fields for end-to-end autonomous driving.
\newblock In \emph{ICCV}, 2021.

\bibitem[Deng et~al.(2009)Deng, Dong, Socher, Li, Li, and Fei-Fei]{imagenet}
Jia Deng, Wei Dong, Richard Socher, Li-Jia Li, Kai Li, and Li~Fei-Fei.
\newblock Imagenet: A large-scale hierarchical image database.
\newblock In \emph{CVPR}, 2009.

\bibitem[Dosovitskiy et~al.(2017)Dosovitskiy, Ros, Codevilla, Lopez, and
  Koltun]{Dosovitskiy17carla}
Alexey Dosovitskiy, German Ros, Felipe Codevilla, Antonio Lopez, and Vladlen
  Koltun.
\newblock {CARLA}: {An} open urban driving simulator.
\newblock In \emph{CoRL}, 2017.

\bibitem[He et~al.(2016)He, Zhang, Ren, and Sun]{he2016resnet}
Kaiming He, Xiangyu Zhang, Shaoqing Ren, and Jian Sun.
\newblock Deep residual learning for image recognition.
\newblock In \emph{CVPR}, 2016.

\bibitem[Jaeger(2021)]{jaeger2021transfuser+}
Bernhard Jaeger.
\newblock Expert drivers for autonomous driving.
\newblock Master's thesis, University of Tübingen, 2021.

\bibitem[Kingma and Ba(2015)]{Adam}
Diederik~P. Kingma and Jimmy Ba.
\newblock Adam: A method for stochastic optimization.
\newblock In \emph{ICLR}, 2015.

\bibitem[Muhammad and Yeasin(2020)]{eigencam}
Mohammed~Bany Muhammad and Mohammed Yeasin.
\newblock Eigen-cam: Class activation map using principal components.
\newblock In \emph{IJCNN}, 2020.

\bibitem[Prakash et~al.(2021)Prakash, Chitta, and
  Geiger]{prakash2021transfuser}
Aditya Prakash, Kashyap Chitta, and Andreas Geiger.
\newblock Multi-modal fusion transformer for end-to-end autonomous driving.
\newblock In \emph{CVPR}, 2021.

\bibitem[Selvaraju et~al.(2017)Selvaraju, Cogswell, Das, Vedantam, Parikh, and
  Batra]{gradcam}
Ramprasaath~R Selvaraju, Michael Cogswell, Abhishek Das, Ramakrishna Vedantam,
  Devi Parikh, and Dhruv Batra.
\newblock Grad-cam: Visual explanations from deep networks via gradient-based
  localization.
\newblock In \emph{ICCV}, 2017.

\bibitem[Toromanoff et~al.(2020)Toromanoff, Wirbel, and
  Moutarde]{toromanoff2020modelfreerl}
Marin Toromanoff, Emilie Wirbel, and Fabien Moutarde.
\newblock End-to-end model-free reinforcement learning for urban driving using
  implicit affordances.
\newblock In \emph{CVPR}, 2020.

\bibitem[Zhang and Ohn-Bar(2021)]{zhang2021lbw}
Jimuyang Zhang and Eshed Ohn-Bar.
\newblock Learning by watching.
\newblock In \emph{CVPR}, 2021.

\bibitem[Zhang et~al.(2021)Zhang, Liniger, Dai, Yu, and
  Van~Gool]{zhang2021roach}
Zhejun Zhang, Alexander Liniger, Dengxin Dai, Fisher Yu, and Luc Van~Gool.
\newblock End-to-end urban driving by imitating a reinforcement learning coach.
\newblock In \emph{ICCV}, 2021.

\end{thebibliography}
}

\appendix



\end{document}


\maketitle

\appendix

In this Supplementary document, we first provide a detailed description of the dataset in Sec.~\ref{sec:dataset}. Implementation and training details are in Sec.~\ref{sec:implementation details}. We show detailed infraction statistics for both leaderboard results and ablation studies, and qualitative results in Sec.~\ref{sec: exp}. Last, we discuss limitations, common failure cases, and possible future directions and potential social impact of our work in Sec.~\ref{sec: discussion}.




\section{Dataset}
\label{sec:dataset}

\subsection{Dataset Collection}
\label{sec:data collection}
We use CARLA 0.9.10.1 for data collection and testing.
We use Roach \citep{zhang2021roach} as the expert to collect data. In order to improve the obstacle avoidance ability of the expert, we additionally add a rule-based vehicle and pedestrian detector adopted from Transfuser \citep{prakash2021transfuser} to avoid possible collisions. Each route is generated randomly with length ranging from 50 meters to 300 meters. We use the scenario configurations provided in \citep{prakash2021transfuser}. We terminate each route if the expert makes a collision or runs a red light. Last few frames for such routes are discarded. The data samples are stored at 2HZ.

\subsection{Dataset Statistics}
\label{sec:dataset statistics}
Detailed statistics for each town and their descriptions are provided in Table~\ref{table:town_detail}. As stated in the main paper, we train on all eight towns for the leaderboard submission. For our ablation experiments, we train on four towns (Town01, Town03, Town04, and Town06) and test on the designed four routes with four different weathers in Town02 and Town05, as does in \citep{chen2022lav}.

\begin{table}[h]
\caption{Detailed statistics of the number of samples, the number of dynamic agents added, and a brief description of each town.}
\label{table:town_detail}
\centering
\scalebox{0.8}{
\begin{tabular}{@{}cccl@{}}
\toprule
Town Name & \#Samples & \#Dynamic Agents & Description                             \\ \midrule
Town01    & 50384                  & 120                      & a basic town with T junction            \\
Town02    & 55943                  & 100                      & similar to Town01 but smaller           \\
Town03    & 42771                  & 120                      & a complex town                          \\
Town04    & 47954                  & 200                      & a highway loop and a small town         \\
Town05    & 53684                  & 120                      & a squared-grid town with multiple lanes \\
Town06    & 48415                   & 150                      & long highways                           \\
Town07    & 51549                  & 110                      & a rural enviroment with narrow roads    \\
Town10    & 59898                  & 120                      & a city with various environments        \\ \bottomrule
\end{tabular}}
\end{table}

\section{Implementation Details}
\label{sec:implementation details}
We use ResNet-34 \citep{he2016resnet} pretrained on ImageNet \citep{imagenet} as the image encoder. The size of the input image is $900 \times 256$ and the FOV of the camera is set as $100\degree$. We choose $K$ being 4 meaning four future steps at 2HZ are predicted for both the trajectory branch and the control branch. Detailed network structure is presented in Table~\ref{table:network_structure}. We follow the same PID setting as \citep{chitta2021neat}, where the PID parameters are exquisitely tuned, \textit{i.e.}, $K_p$ = 5.0, $K_i$ = 0.5, $K_d$ = 1.0 for the longitudinal PID controller and $K_p$ = 0.75, $K_i$ = 0.75, $K_d$ = 0.3 for the lateral PID controller. The weights for different loss terms are as follows: $\lambda_{F} = 0.05$, $\lambda_{traj} = 1$, $\lambda_{ctl} = 1$, $\lambda_{aux} = 0.05$, and $0.001$ for speed and value regression respectively.
For all experiments, we train \algname on 4 GeForce RTX 3090 GPUs. We use the Adam optimizer \citep{Adam} with a learning rate of $1\times 10^{-4}$ and weight decay of $1\times10^{-7}$ for all experiments. We train all models with batch size 128 for 60 epochs, and the learning rate is reduced by a factor of 2 after 30 epochs.

In the situation based fusion scheme, we choose whether the vehicle is turning as the criterion of the  $situation$. Specifically, we calculate the absolute values of steer actions within the past 1 second. If half of them are larger than 0.1, we assume the vehicle is turning so the $situation$ is $control\ specialized$, otherwise $trajectory\ specialized$.
For the online CARLA Leaderboard \citep{carlaleaderboard} submission, we use an asymmetric fusion scheme.
%
If the $situation$ is $trajectory\ specialized$, we set $\alpha = 0.5$, and $\alpha = 0$ when it is $control\ specialized$. We take the maximum of the $brake$ control instead of taking the average. For the ensemble submission TCP-Ens, we also take the maximum of $brake$ value from different models and take the average for $steer$ and $throttle$.

\section{Experiments}
\label{sec: exp}

\subsection{Validation Protocol Details}

We use the same validation routes as LAV \citep{chen2022lav}. This includes 4 routes in total, 2 from Town02 and 05 each. Each route is tested under 4 different weathers (ClearNoon, CloudySunset, SoftRainDawn, HardRainNight) and is repeated for 3 times, resulting in 48 routes in total. Random scenarios are added from the official CARLA leaderboard repo (all\_towns\_traffic\_scenarios\_public.json). The time-limit for agent blocking is reduced from 300 seconds to 60 seconds to save time.

\subsection{Detailed Infractions Statistics}

In this part, we report detailed infraction statistics
of the methods on CARLA Leaderboard in Table~\ref{table:comparesota_detail}, and statistics of our ablation experiments in Table~\ref{table:ablation_detail} and Table~\ref{table:mtlvsensemble_detail}.

\begin{table}[t!]
\centering
\caption{Detailed statistics of the evaluation on the public CARLA Leaderboard \citep{carlaleaderboard} (accessed in May 2022). Driving Score, Route Completion, and Infraction Penalty are higher the better. For other metrics, lower values are desired. The collisions, infractions, and agent blocked related metrics are given as the number of events per kilometer.    Our method outperforms other methods by a large margin in terms of Driving Score and Route Completion. We also have the best scores for metrics of collisions vehicle, collisions pedestrian, collisions layout, and off-road infractions among all methods.}
\label{table:comparesota_detail}
\scalebox{0.7}{
\begin{tabular}{@{}llccccccccc@{}}
\toprule
Rank & Method & \begin{tabular}[c]{@{}c@{}}Driving\\ Score\end{tabular} & \begin{tabular}[c]{@{}c@{}}Route\\ Completion\end{tabular} & \begin{tabular}[c]{@{}c@{}}Infraction\\ Penalty\end{tabular} & \begin{tabular}[c]{@{}c@{}}Collisions\\ Vehicle\end{tabular} & \begin{tabular}[c]{@{}c@{}}Collisions\\ Pedestrian\end{tabular} & \begin{tabular}[c]{@{}c@{}}Collisions\\ Layout\end{tabular} & \begin{tabular}[c]{@{}c@{}}Red light\\ Infractions\end{tabular} & \begin{tabular}[c]{@{}c@{}}Off-road\\ Infractions\end{tabular} & \begin{tabular}[c]{@{}c@{}}Agent\\ Blocked\end{tabular} \\ \midrule
1& \textbf{TCP-Ens} (ours) & \textbf{75.137} & 85.629 & \textbf{0.873} & 0.316 & \textbf{0.000} & \textbf{0.000} & 0.089 & 0.038 & 0.537 \\
1& \textbf{TCP} (ours) & 69.714 & 82.962 & 0.851 & \textbf{0.220} & 0.006 & 0.034 & 0.083 & \textbf{0.017} & 0.564\\
1& \textbf{TCP-SB} (ours) & 68.695 & 82.957 & 0.833 & 0.250 & \textbf{0.000} & 0.111 & 0.066 & 0.026 & 0.528 \\
\midrule
2& LAV \citep{chen2022lav}& 61.846 & \textbf{94.459}& 0.640 & 0.696 & 0.038& 0.017& 0.166 & 0.252 & \textbf{0.104} \\
3 & Transfuser& 61.181 & 86.694 & 0.714 & 0.814 & 0.036& 0.007& \textbf{0.046} & 0.228 & 0.428 \\
4 & Latent Transfuser& 45.029 & 75.366 & 0.618 & 1.259 & 0.034& 0.098& 0.102 & 0.288 & 0.757 \\
5 &GRIAD \citep{chekroun2021gri}& 36.787 & 61.855 & 0.597 & 2.772 & \textbf{0.000}& 0.407& 0.484 & 1.388 & 0.842 \\
6 &Transfuser+ \citep{jaeger2021transfuser+}& 34.577 & 69.841 & 0.562 & 0.703 & 0.045 & 0.025 & 0.750 & 0.185 & 2.406 \\
7 &WoR \citep{chen2021wor}& 31.370 & 57.647 & 0.557 & 1.346 & 0.606 & 1.017 & 0.791 & 0.963 & 0.473 \\
8 &MaRLn \citep{toromanoff2020modelfreerl}& 24.980 & 46.968 & 0.518 & 2.329 & \textbf{0.000} & 2.472 & 0.550 & 1.823 & 0.936 \\
9&NEAT \citep{chitta2021neat}& 
21.832 & 41.707 & 0.650 &0.742 & 0.042 & 0.617 & 0.700 & 2.680 & 5.225 \\
\bottomrule
\end{tabular}
}
\end{table}

\begin{table}[t!]
\centering
\caption{Detailed infraction statistics of the ablation on the effectiveness of the trajectory-guided multi-step control prediction design.}
\label{table:ablation_detail}
\scalebox{0.72}{
\begin{tabular}{@{}lccccccccc@{}}
\toprule
Exp. & \begin{tabular}[c]{@{}c@{}}Driving\\ Score\end{tabular} & \begin{tabular}[c]{@{}c@{}}Route\\ Completion\end{tabular} & \begin{tabular}[c]{@{}c@{}}Infraction\\ Penalty\end{tabular} & \begin{tabular}[c]{@{}c@{}}Collisions\\ Vehicle\end{tabular} & \begin{tabular}[c]{@{}c@{}}Collisions\\ Pedestrian\end{tabular} & \begin{tabular}[c]{@{}c@{}}Collisions\\ Layout\end{tabular} & \begin{tabular}[c]{@{}c@{}}Red light\\ Infractions\end{tabular} & \begin{tabular}[c]{@{}c@{}}Off-road\\ Infraction\end{tabular} & \begin{tabular}[c]{@{}c@{}}Agent\\ Blocked\end{tabular} \\ \midrule
Control& 32.45$\pm$2.23 & 76.54$\pm$3.22 & 0.45$\pm$0.03 & 1.24$\pm$0.06 & 0.00$\pm$0.00 & 0.23$\pm$0.09 & 0.18$\pm$0.05 & 0.59$\pm$0.06 & 0.41$\pm$0.11 \\
+ traj-task& 34.98$\pm$1.96 &81.32$\pm$5.50 &0.49$\pm$0.05 & 1.39$\pm$0.15 & 0.00$\pm$0.00 &0.15$\pm$0.07 &0.11$\pm$0.04 &0.39$\pm$0.04 & 0.38$\pm$0.10 \\
+ temporal&42.87$\pm$4.77 &\textbf{87.51}$\pm$3.63 &0.49$\pm$0.07 &1.14$\pm$0.25 &0.00$\pm$0.00 &0.20$\pm$0.07 &0.18$\pm$0.04 & 0.18$\pm$0.05 &0.22$\pm$0.03 \\
+ traj-attn&46.08$\pm$3.47 &84.95$\pm$1.84 &0.56$\pm$0.03 &0.90$\pm$0.20 &0.00$\pm$0.00 &\textbf{0.04}$\pm$0.06 &0.14$\pm$0.07 &0.54$\pm$0.06 &0.29$\pm$0.08 \\
+ fusion& \textbf{57.01}$\pm$1.88 &85.27$\pm$1.20 & \textbf{0.67}$\pm$0.01 &\textbf{0.37}$\pm$0.10 &0.00$\pm$0.00 & 0.08$\pm$0.03 & \textbf{0.10}$\pm$0.03 & \textbf{0.14}$\pm$0.06 &\textbf{0.20}$\pm$0.03 \\
\bottomrule
\end{tabular}}
\end{table}

\begin{table}[t!]
\centering
\caption{Detailed infraction statistics of the experiments of the comparison between MTL and ensemble methods.}
\label{table:mtlvsensemble_detail}
\scalebox{0.72}{
\begin{tabular}{@{}lccccccccc@{}}
\toprule
Exp. & \begin{tabular}[c]{@{}c@{}}Driving\\ Score\end{tabular} & \begin{tabular}[c]{@{}c@{}}Route\\ Completion\end{tabular} & \begin{tabular}[c]{@{}c@{}}Infraction\\ Penalty\end{tabular} & \begin{tabular}[c]{@{}c@{}}Collisions\\ Vehicle\end{tabular} & \begin{tabular}[c]{@{}c@{}}Collisions\\ Pedestrian\end{tabular} & \begin{tabular}[c]{@{}c@{}}Collisions\\ Layout\end{tabular} & \begin{tabular}[c]{@{}c@{}}Red light\\ Infractions\end{tabular} & \begin{tabular}[c]{@{}c@{}}Off-road\\ Infraction\end{tabular} & \begin{tabular}[c]{@{}c@{}}Agent\\ Blocked\end{tabular} \\ \midrule
Ensemble&45.03$\pm$1.28 &79.30$\pm$5.13 &0.59$\pm$0.04 &0.62$\pm$0.09 &0.00$\pm$0.00 &0.22$\pm$0.03 &0.22$\pm$0.07 &0.28$\pm$0.03 &0.35$\pm$0.10 \\
MTL&48.27$\pm$0.58 & 81.62$\pm$2.74 &0.60$\pm$0.02 &0.51$\pm$0.07 &0.00$\pm$0.00 & 0.26$\pm$0.06 &\textbf{0.06}$\pm$0.05 &0.36$\pm$0.03 & 0.28$\pm$0.09 \\
TCP-SB&52.46$\pm$4.66 &83.94$\pm$3.75 &0.64$\pm$0.04 &0.53$\pm$0.14 &0.00$\pm$0.00 &0.08$\pm$0.03 &0.13$\pm$0.09 &\textbf{0.06}$\pm$0.00 &0.29$\pm$0.04 \\
TCP& 57.01$\pm$1.88 &85.27$\pm$1.20 & 0.67$\pm$0.01 &\textbf{0.37}$\pm$0.10 &0.00$\pm$0.00 & 0.08$\pm$0.03 & 0.10$\pm$0.03 & 0.14$\pm$0.06 &\textbf{0.20}$\pm$0.03 \\
TCP-Ens&\textbf{59.09}$\pm$3.66 &\textbf{87.02}$\pm$2.02 &\textbf{0.70}$\pm$0.03 &0.41$\pm$0.19 &0.00$\pm$0.00 &\textbf{0.00}$\pm$0.00 &0.10$\pm$0.13 &0.18$\pm$0.05 &0.27$\pm$0.06 \\
\bottomrule
\end{tabular}}
\end{table}

\subsection{Qualitative Results}
\begin{figure}[t!]
    \centering
    \includegraphics[width=1.0\textwidth]{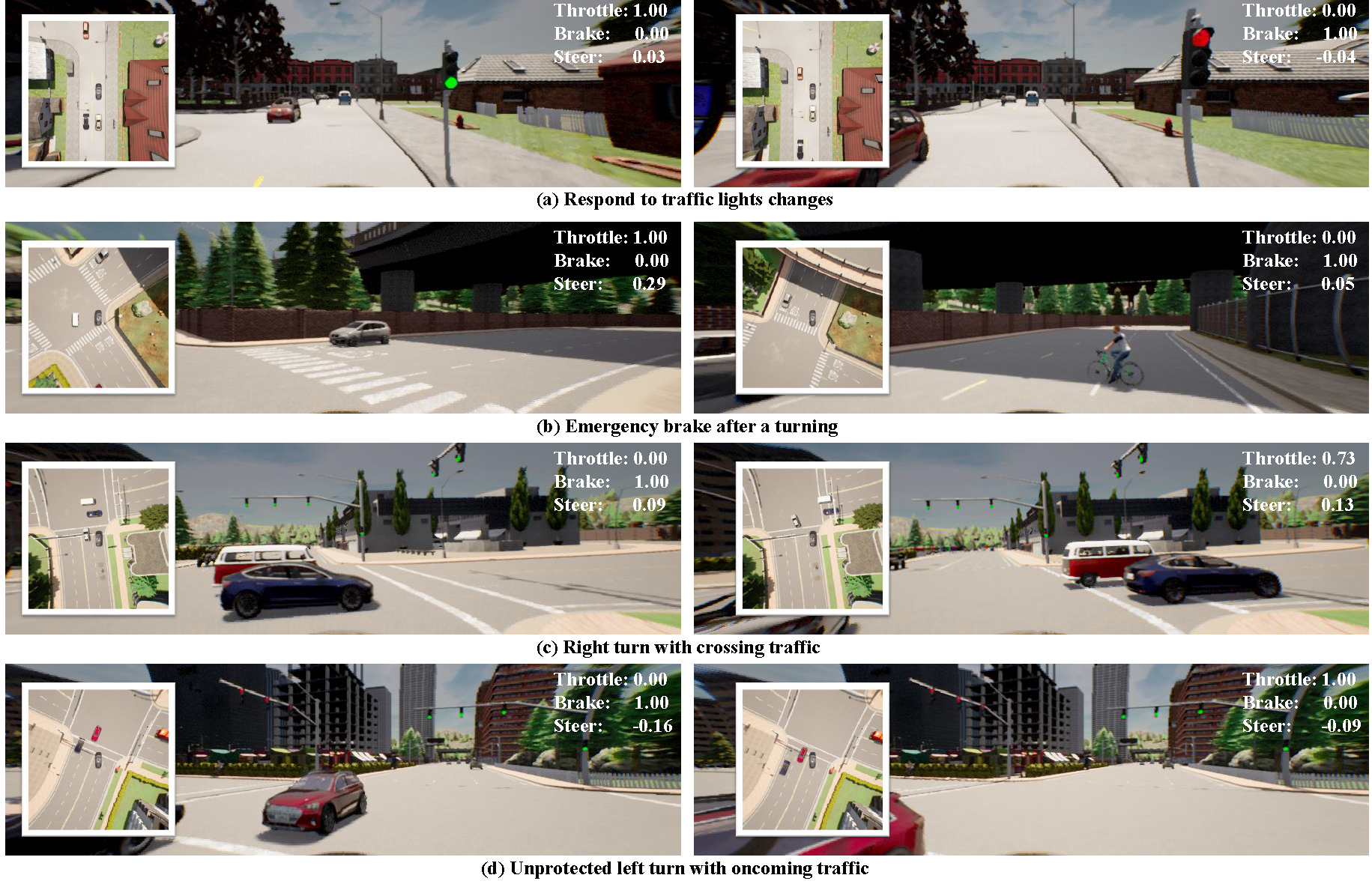}
    \caption{Examples of our agent performing well under different challenging scenarios.}
    \label{fig:cases}
\end{figure}

We show cases of our method performing well in different challenging scenarios in Fig.~\ref{fig:cases}. In the first case, the autonomous agent successfully reacts to the changing of the traffic light in time at the crossing.
In the second case, a cyclist suddenly runs across the road right after the ego vehicle has made a right turn, and our agent makes an emergency brake in time, avoiding a collision.
In the third case, the ego vehicle is making a right turn while there are other vehicles crossing. It stops and waits for the crossing vehicles to pass and then continues to make the turn.
In the last case, our agent is performing an unprotected left turn with oncoming traffic, and it successfully negotiates with the oncoming vehicle.

\begin{figure}[t!]
    \centering
    \includegraphics[width=1.0\textwidth]{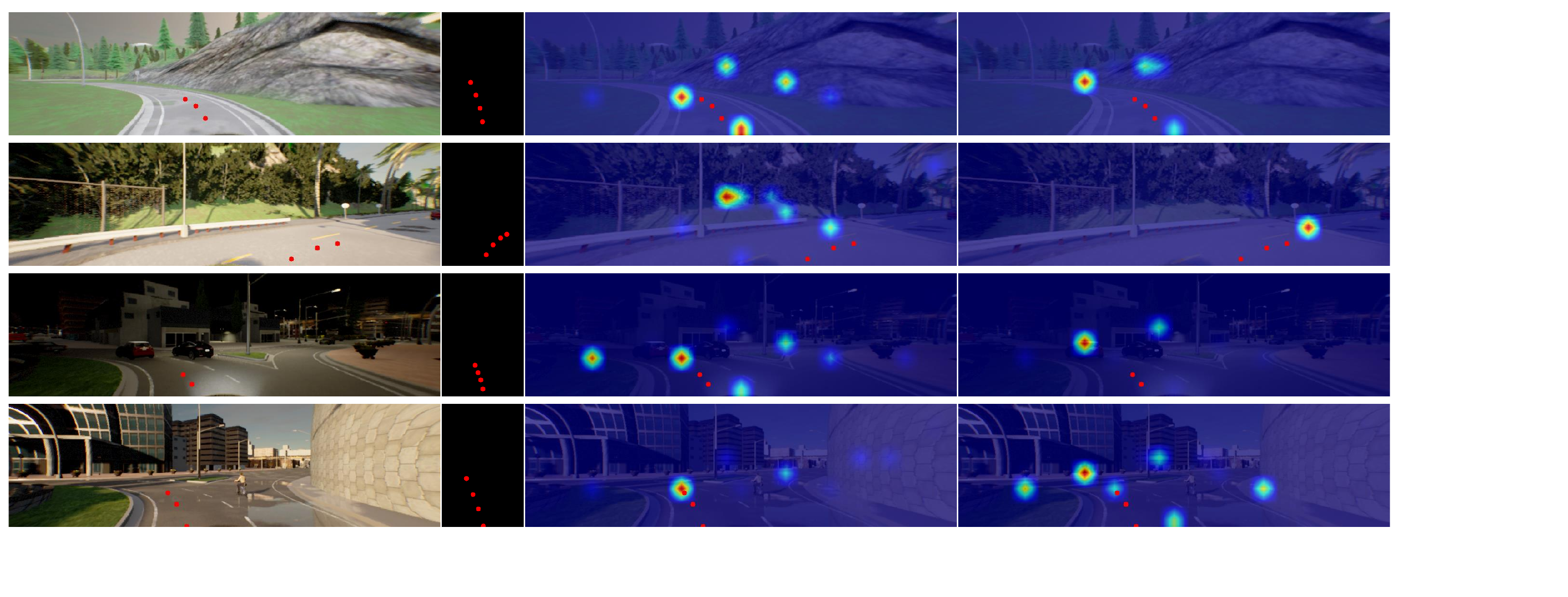}
    \caption{More examples of  trajectory-guided attention maps. In each case (row), from the left to right we show that the input image with the predicted trajectory (the first waypoint is projected out of the image), the predicted trajectory in the top-down view, the attention map $\rm{\bf w}_1$, the attention map $\rm{\bf w}_3$.}
    \label{fig:att_supp}
\end{figure}

\begin{figure}
    \centering
    \includegraphics[scale=0.42]{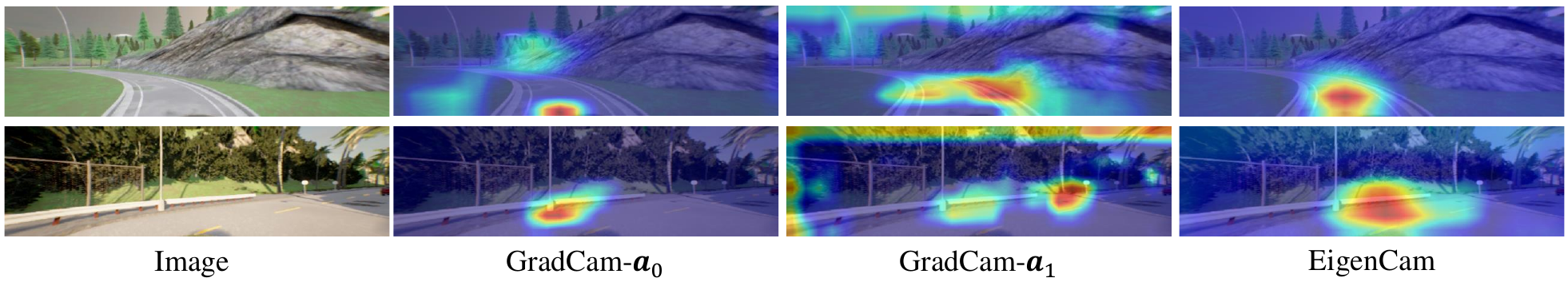}
    \caption{Visualization examples of GradCam \citep{gradcam} and EigenCam \citep{eigencam}. From left to right: original image, GradCam heat-map for current action prediction, GradCam heat-map for future action prediction, EigenCam heat-map for the image feature map.}
    \label{fig:cam}
\end{figure}

More visualization examples of the trajectory-guided attention maps are provided in Fig.~\ref{fig:att_supp}.
We also show the  GradCam~\citep{gradcam} and EigenCam~\citep{eigencam} visualization of two examples for Control-Only model with multi-step prediction scheme in Fig.~\ref{fig:cam}. 
For the GradCam visualization, we set the target (which is needed to be maximized during the calculation of GradCam) be the negative action loss for the current and future action prediction.
Note that GradCam visualizes the regions of the input image that are \textbf{important} for predictions by calculating gradients to maximize the target. It \textbf{does not} indicate that the model does focus or well capture the region highlighted by GradCam.
As shown in Fig.~\ref{fig:cam}, the GradCam heat-map for current action prediction (${\bf a}_0$) focuses on regions close to the current location of the ego vehicle, while the heat-map for future prediction (${\bf a}_1$) focuses on regions further.
This indicates that predicting future actions does need to focus on further regions.

However, Control-Only model with multi-step action prediction only aggregates the image feature map once by global average. The pooled feature is then used to predict actions of all time steps.
Therefore, it is \textbf{not} realistic to highlight corresponding important regions for each step.
We use the EigenCam to visualize the 2D image feature map. It is a gradient-free visualization method to directly calculate the heat-map by projecting the feature map to eigen-vectors.
As shown in the last column in Fig.~\ref{fig:cam}, the highlighted region of the 2D image feature map only spans a single area, which is not informative enough for multi-step predictions.
It verifies the necessity of re-aggregating the image information with different highlighted regions for each future step, as what we did in Fig.~\ref{fig:att_supp}.

\section{Discussion}
\label{sec: discussion}
\subsection{Limitations and Future Work}

\begin{figure}
    \centering
    \includegraphics[scale=0.4]{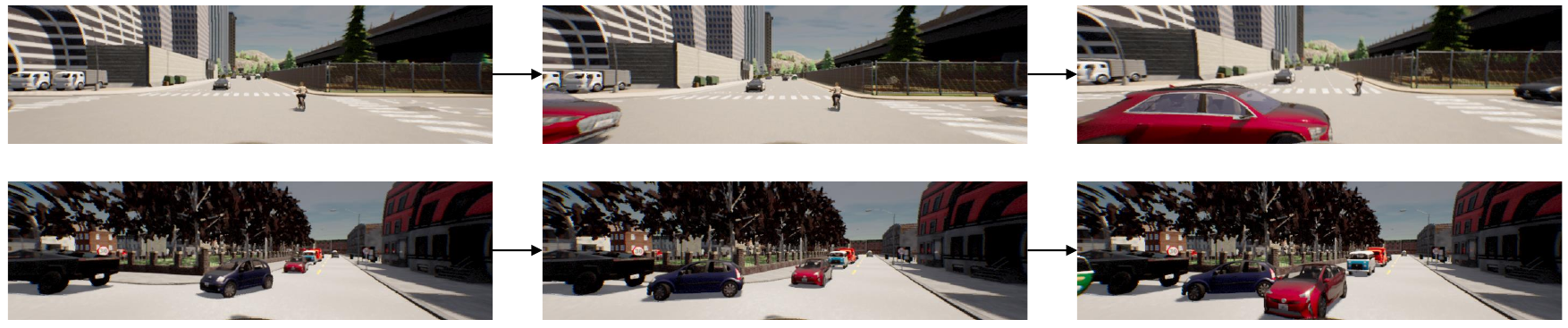}
    \caption{Examples of two failure cases. Top row: the red vehicle runs into the ego path with a high speed, and the ego vehicle fails to take an emergent brake.
    Down row: The ego agent is waiting for a left turn but occupies part of the opposite lane, causing a block.}
    \label{fig:failure_case}
\end{figure}

\subsubsection{Failure Cases and Future Work Directions}
Our work mainly focuses on combining the two output forms of end-to-end autonomous driving, \textit{i.e.}, trajectory planning and direct control. A detailed and elaborate situation based fusion scheme is based on rules which may require a large number of experiments and specific prior knowledge. A more general or a learning-based adaptive fusion scheme may be a possible future direction.

We also discuss two typical failure cases of \algname in Fig.~\ref{fig:failure_case}. The first scenario happens when other vehicles initially outside the ego agent's front view rushes into the path with a high speed. It causes a delayed collision when an emergent braking fails. It is because of the limited view of our single camera, hence a straightforward future direction is to add multi-view cameras or a LiDAR input to our agent. Another kind of failures is that the ego agent fails to predict the possible trajectory of other vehicles, resulting in blocking or collisions. Thus explicitly making trajectory predictions of other vehicles and combining it with our trajectory branch is also an interesting direction to further boost the ability of generalization as demonstrated in LAV \citep{chen2022lav} and LBW \citep{zhang2021lbw}.

\subsection{Broader Impact}
We explore the limitations and advantages of the two conventional output paradigms for end-to-end autonomous driving, present \algname which achieves state-of-the-art performance on the public closed-loop benchmark to push the boundary of the problem. We aim to bring together the two branches of research in this field and provide a unified framework to combine their possible advantages. Our work provides a simple yet effective framework, based on which, new models and techniques can be conveniently integrated and transparently compared.
Despite such improvement, we fully understand that our work is by no means perfect and still has many challenges when it comes to real-world application. Our model is trained and tested in the simulator, directly deploying it in the real world will lead to possible traffic accidents which may cause negative societal impacts.

\section{License of Assets}
CARLA \citep{Dosovitskiy17carla} is an open-source simulator which is under the MIT license and its assets are under the CC-BY license. We integrate part of the official code of Roach \citep{zhang2021roach} which is under the CC-BY-NC 4.0 license into our codebase. The pretrained ResNet model is under the MIT license.

The source code and training data for our work will be publicly available once accepted and they are under the CC-BY-NC 4.0 license.

\begin{table}[]
\centering
\caption{Detailed network structure of our TCP model.}
\begin{tabular}{@{}lccc@{}}
\toprule
Layer Type  & \# of Filters     & Activation Function & \# \\ \midrule
\multicolumn{4}{c}{Image Encoder}                         \\ \midrule
ResNet-34    &                   &                     &    \\ \midrule
\multicolumn{4}{c}{Measurement Encoder}                   \\ \midrule
FC          & 128               & ReLU                & $\times$ 2  \\ \midrule
\multicolumn{4}{c}{Join\_ctrl}                            \\ \midrule
FC          & 512               & ReLU                & $\times$ 2  \\
FC          & 256               & ReLU                & $\times$ 1  \\ \midrule
\multicolumn{4}{c}{Join\_traj}                            \\ \midrule
FC          & 512               & ReLU                & $\times$ 2  \\
FC          & 256               & ReLU                & $\times$ 1  \\ \midrule
\multicolumn{4}{c}{Speed Head}                            \\ \midrule
FC          & 256               & ReLU                & $\times$ 2  \\
FC          & 1                 & ReLU                & $\times$ 1  \\ \midrule
\multicolumn{4}{c}{Value Head $\times 2$ traj+ctrl}                  \\ \midrule
FC          & 256               & ReLU                & $\times$ 2  \\
FC          & 1                 & ReLU                & $\times$ 1  \\ \midrule
\multicolumn{4}{c}{Temporal Module}                       \\ \midrule
GRU\_cell   & hidden size = 256 &                     & $\times$ 1  \\
FC (output) & 256               & ReLU                & $\times$ 2  \\ \midrule
\multicolumn{4}{c}{Control Policy Head}                   \\ \midrule
FC          & 256               & ReLU                & $\times$ 2  \\
FC (alpha)  & 2                 & Softplus            & $\times$ 1  \\
FC (beta)   & 2                 & Softplus            & $\times$ 1  \\ \midrule
\multicolumn{4}{c}{Traj Policy Head}                       \\ \midrule
GRU\_cell   & hidden size = 256 &                     & $\times$ 1  \\
FC (output) & 2                 &                     & $\times$ 2  \\ \midrule
\multicolumn{4}{c}{Init Att.}                              \\ \midrule
FC          & 256               & ReLU                & $\times$ 1  \\
FC          & 29*8              & Softmax             & $\times$ 1  \\ \midrule
\multicolumn{4}{c}{Traj Guided Att.}                       \\ \midrule
FC          & 256               & ReLU                & $\times$ 1  \\
FC          & 29*8              & Softmax             & $\times$ 1  \\ \midrule
\multicolumn{4}{c}{Merge (merge the re-aggregated image feature and hidden state)}                                  \\ \midrule
FC          & 512               & ReLU                & $\times$ 1  \\
FC          & 256               & ReLU                & $\times$ 1  \\ \bottomrule
\end{tabular}
\label{table:network_structure}
\end{table}

\bibliographystyle{plainnat}
{
\small
\bibliography{egbib}
}